\documentclass{article}

\usepackage{microtype}
\usepackage{graphicx}
\usepackage{caption}
\usepackage{subcaption}
\usepackage{booktabs}
\usepackage{hyperref}
\usepackage[utf8]{inputenc}
\usepackage[T1]{fontenc}   
\usepackage{url}           
\usepackage{amsfonts}      
\usepackage{nicefrac}      
\usepackage{microtype}     
\usepackage{mypackage}
\usepackage{algorithm}
\usepackage{algpseudocode}
\usepackage{amssymb}
\usepackage{xcolor}
\usepackage{cleveref}
\usepackage{caption}
\graphicspath{{./}}

\usepackage{color}
\usepackage{epigraph}
\usepackage[accepted]{icml2020}
\icmltitlerunning{GraphOpt: Learning Optimization Models of Graph Formation}
\begin{document}

\twocolumn[
\icmltitle{GraphOpt: Learning Optimization Models of Graph Formation}
\icmlsetsymbol{equal}{*}

\begin{icmlauthorlist}
\icmlauthor{Rakshit Trivedi}{gt}
\icmlauthor{Jiachen Yang}{gt}
\icmlauthor{Hongyuan Zha}{chk,gt}
\end{icmlauthorlist}

\icmlaffiliation{gt}{Georgia Institute of Technology, USA}
\icmlaffiliation{chk}{Institute for Data and Decision Analytics, the Chinese University of Hong Kong, Shenzhen}

\icmlcorrespondingauthor{Rakshit Trivedi}{rstrivedi@gatech.edu}

\icmlkeywords{Graph formation, Optimization Mechanisms, Inverse Reinforcement Learning}

\vskip 0.3in
]

\printAffiliationsAndNotice{}

\begin{abstract}

Formation mechanisms are fundamental to the study of complex networks, but learning them  from observations is challenging. In real-world domains, one often has access only to the final constructed graph, instead of the full construction process, and observed graphs exhibit complex structural properties. In this work, we propose GraphOpt, an end-to-end framework that jointly learns an implicit model of graph structure formation and discovers an \textit{underlying optimization mechanism} in the form of a \textit{latent} objective function. 
The learned objective can serve as an explanation for the observed graph properties, thereby lending itself to transfer across different graphs within a domain. GraphOpt poses link formation in graphs as a sequential decision-making process and solves it using maximum entropy inverse reinforcement learning algorithm. Further, it employs a novel continuous \textit{latent action space} that aids scalability. 
Empirically, we demonstrate that GraphOpt discovers a latent objective transferable across graphs with different characteristics. GraphOpt also learns a robust stochastic policy that achieves competitive link prediction performance without being explicitly trained on this task and further enables construction of graphs with properties similar to those of the observed graph.

\end{abstract}

\section{Introduction and Related Work}

Learning generative mechanisms of graph-structured data is an important approach to build graph constructors for data augmentation~\cite{ChaFal06} and inference modules for downstream network analysis and prediction tasks. 
Such models have wide-ranging applications in several domains spanning recommendation systems~\cite{ZhaYaoSun18}, biological networks~\cite{SinLio19}, knowledge graphs~\cite{XiaHuaZhu16}, social networks~\cite{LiuCheLiLia16} and many more. Formation mechanisms play a fundamental role in driving the generative process of structures observed in the real-world~\cite{barabasi2016network}. Modeling these mechanisms is important as it could facilitate synthesis of novel graph structures for various subsequent studies such as analysing disease progression.
Moreover, access to these mechanisms could help to build intelligent systems that \textit{generalize} beyond the task of structure generation and support \textit{transfer} to graphs beyond available observations. 
However, in real-world domains, such formation mechanisms are often unknown and learning them from data is challenging due to: limited or no access to the construction process (one often observes only the final graph); complex discrete nature of graph and rare availability of large collection of graph samples for learning. 

\begin{figure}[t]
\centering
  \vspace{-3mm}
  \resizebox{0.25\textwidth}{!}{
  \begin{tabular}{c}
  \includegraphics[width=0.5\textwidth]{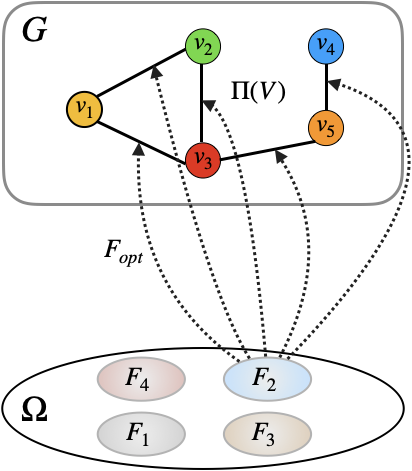}
  \end{tabular}}
  \caption{
  $\Omega$ is set of latent objective functions $\lbrace \mathcal{F}_i \rbrace$, any of which could lead to observed graph $\mathcal{G}$ when optimised.
  Our goal is to discover one such latent objective $\mathcal{F}_{\text{opt}}$ that could serve as an explanation of the observed graph properties, and optimise it to learn a graph construction procedure $\Pi$ such that $\Pi(\mathcal{V})$, given node set $\mathcal{V}$, mimics the network patterns observed in $\mathcal{G}$.
  While $\mathcal{F}_{\text{opt}}$ may not match the unknown ground truth mechanism when one exists, it can produce an accurate $\Pi$ and hence can be operationally equivalent to the true mechanism.}
  \label{fig:goptoverview}
  \vspace{-0.3cm}
\end{figure}

In this paper, we investigate the novel problem setting of discovering an underlying formation mechanism of the observed graph structure. Concretely, we propose \textbf{GraphOpt}, an end-to-end learning framework that jointly learns the forward model of graph construction and solves the inverse problem of discovering an \textit{underlying optimization mechanism}, in the form of a \textit{latent} objective function, that serves as an explanation for the existence of the observed graph structure. 
From a broader perspective of network science, GraphOpt naturally aligns more with the "optimization" viewpoint of graph formation~\cite{newman2010networks,papadopoulos2012popularity,barabasi2016network}---link formation is viewed as the outcome of an underlying optimization mechanism whereby decisions are based on current global state of the network.
For instance, links in transportation networks appear as a result of optimizing some underlying cost function~\cite{AlbBar02}.
In contrast, most existing learning approaches for graphs are implicitly rooted in the "probabilistic" viewpoint,
which models link formation in networks as random events that depend largely on local structural properties~\cite{kumar2000web,vazquez2003modeling}. \Cref{fig:goptoverview} provides an overview of our proposition.

Formally, GraphOpt is realised as an efficient maximum entropy based reinforcement learning~\citep{haarnoja2018soft,ZieMaaBagdey08} framework that models graph formation as a sequential decision-making process. It trains a novel structured policy network such that the learned stochastic policy constructs edges in a sequential  manner to produce a graph with minimal deviation in a set of graph properties from an observed graph. This policy network uses a graph neural network to capture the complex information of a partially constructed graph into a continuous state representation. As the true graph formation objective function is unknown, the policy optimizes a latent reward function learned via inverse reinforcement learning (IRL)~\citep{ZieMaaBagdey08,finn2016guided}, which amounts to learning an optimization-based model of graph formation. Further, we propose a novel \textit{continuous latent action space} that is independent of the size of graph, thereby allowing GraphOpt to learn over large graphs. 

Traditional generative approaches include explicit probabilistic models that are carefully hand-designed to incorporate assumptions on structural properties~\cite{RobPatKalLus07, LesKleFal07, AirBleFieXin09,LesChaKleFal10,ErdRen59}.
Such intuitive model specifications produce graphs that often exhibit disagreement with real-world graphs~\cite{BroClau18, DonJohXuCha17}.
Recent advances in deep generative models for graphs~\cite{li2018learning,SimKom18,you2018graphrnn,kipf2016variational} address this issue by directly learning from data to be able to mimic the observed properties and produce realistic graphs. However, these techniques are either limited to learn over small graphs or require a large collection of graphs from the same distribution to achieve a desired fidelity, both of 
which pose great limitations on learning over real-world graphs. Further, the above techniques only facilitate graph generation but does not directly allow downstream inference tasks, which further limits their usability.
A recently proposed deep generative model NetGAN~\cite{BojShcZugCun18}, resembles GraphOpt in being implicit model for graph construction, however, the two approaches are fundamentally different as NetGAN builds a probabilistic model of random walks over graphs and avoids learning an objective function which stands in contrast to our optimization-based framework. 

We perform extensive experiments on graphs with varying properties to gauge the efficacy of GraphOpt on the following measures: 
(i) Can GraphOpt discover a useful and transferable latent objective for a given domain? (ii) How well does GraphOpt’s construction policy generalize to downstream inference tasks? (iii) Can GraphOpt serve as an effective graph constructor useful to synthesize new graphs that exhibit structural patterns similar to the ones found in an observed graph?
We comprehensively answer all three questions in the positive via experiments demonstrating effective transfer in the domain of citation graphs; competitive link prediction performance against dedicated baselines demonstrating compelling generalization properties; and  consistently superior performance on graph construction experiments against strong baselines that learn from single input graph. 
We discuss more related works in \Cref{app:relatedworks}.

\vspace{-0.3cm}
\section{Proposed Approach: GraphOpt}

We first elaborate on the optimization viewpoint of graph formation and how it motivates our modeling approach. Next, we formally define the problem we tackle and define the corresponding sequential decision-making process. Finally, we present architecture details of the GraphOpt framework. 

\subsection{Optimization Models of Graph Formation}

A reasonable graph formation model can help to determine how networks come into existence, which can be fundamentally important in various applications where the network structure often influences decision making~\cite{PauRicTam18}. 
While networks in certain domains (e.g. transportation~\cite{AlbBar02}) can be explained by an underlying optimization mechanism with a known functional form, most general networks often exhibit properties that have given rise to long-standing debates in the network science community on the true mechanisms underlying their emergence~\cite{barabasi2012network}. 
For instance, power laws observed in social and biological networks can be explained by the probabilistic model of preferential attachment~\cite{barabasi1999emergence}, but they can also be the result of an underlying optimization process in which  nodes optimize between popularity and similarity when forming connections~\cite{papadopoulos2012popularity,d2007emergence,fabrikant2002heuristically}. For complex networks, the functional form of the objective being optimized in this process is often unknown or difficult to specify in closed form.

In this work, we rigorously investigate the optimization viewpoint and its implications on developing learning approaches for graphs.
As the true underlying mechanisms are unknown, we design an algorithm that \textit{discovers} a latent objective that is operationally equivalent to the true mechanism, in the sense that the discovered objective, when optimised, enables a suitable graph construction procedure to produce graphs with similar properties as the observed one.
As both the construction procedure and the objective depend on the global information, our approach is naturally aligned with the optimization viewpoint of graph formation.

\subsection{Problem Definition}

Given a graph $\mathcal{G} = (\mathcal{V}, \mathcal{E})$, we propose a graph formation model in which the optimization of some latent objective function $\Fcal_{\text{opt}} \colon \Gcal \rightarrow \Rbb$ drives the formation of edges in $\Ecal$.
$\Fcal_{\text{opt}}$ may correspond to any domain-specific or generic graph property but is unknown in general.
Our primary aim is to learn a graph construction procedure $\Pi^*$ and discover a latent objective $\Fcal_{\text{opt}}$, such that the optimization of $\Fcal_{\text{opt}}$ by $\Pi^*$ leads to the construction of a graph $\Gcal' = \Pi^*(\Vcal)$ with  structural properties similar to the observed graph $\Gcal$, given node set $\mathcal{V}$ and initially empty edge set $\Ecal_0$.
Learning the construction procedure is formalized as:
\begin{equation}
\label{eq:obj-argmax}
    \Pi^* = \argmax_\Pi \mathcal{F}_{\rm opt}(\Pi(\mathcal{V})),
\end{equation}

$\mathcal{F}_{\rm opt}$ is often unknown as we only observe the final graph.
To this end, we draw inspiration from inverse reinforcement learning~\cite{ng2000} and formulate our objective as the following minmax optimization problem:
\begin{equation}
\label{eq:obj-minmax}
\begin{split}
    \Pi^* &= \argmin_{\Pi} \max_{\mathcal{F}}\left[ \mathcal{F}(\mathcal{G}) - \mathcal{F}(\Pi(\mathcal{V})) \right] \\
    \mathcal{F}_{\rm opt} &= \argmax_{\mathcal{F}}\left[ \mathcal{F}(\mathcal{G}) - \mathcal{F}(\Pi^*(\mathcal{V})) \right]
\end{split}
\end{equation}

where the goal is to learn jointly: (i) a latent objective $\Fcal$, defined via a reward function, that assigns higher value to $\Gcal$ than to all other graphs with different structural properties;
(ii) a construction procedure $\Pi$, defined as the sequential execution of a policy, that 
constructs $\Gcal'$ with minimal difference from $\Gcal$ in structural properties measured by $\Fcal$.

\subsection{Graph Formation as a Markov Decision Process}

The graph formation mechanism is the central focus of our work.
Formation of real-world graphs in general is not confined to result in a connected graph. Therefore, we propose a mechanism for link formation without this constraint.
Let $\mathcal{G} = (\mathcal{V}, \mathcal{E}, \mathcal{Y}, \mathcal{X})$ denote a graph, where $\mathcal{V}$ is the set of nodes, $\mathcal{E}$ is the set of edges, $\mathcal{Y}$ is the set of edge types and $\mathcal{X}$ is the set of node features.
We define a Graph Formation Markov decision process (GF-MDP) $\mathcal{M} = (\mathcal{S}, \mathcal{A}, \mathcal{R}, P)$ as follows:

{\bf State} $s_t \in \Scal$.
The state of the environment $s_t$ at time $t$ is the partially constructed graph $G_t = (\mathcal{V},\mathcal{E}_t, \mathcal{Y}, \mathcal{X})$. 
Initial state $s_0 = G_0$ is a graph with all nodes but no edges, i.e. $\mathcal{E}_t = \emptyset$. 
Node features $\mathcal{X}$ and edge types $\mathcal{Y}$ are optional. 
This definition is sufficient to describe the graph at any time $t$ and allows for sequential construction of edges without enforcing connectivity.
For ease of exposition and w.l.o.g., we let $ s_t = (\mathcal{V},\mathcal{E}_t)$ represent a state in this paper.

{\bf Action} $a_t \in \Acal$. 
Each step in procedure $\Pi$ involves the creation of an edge between two nodes in $\mathcal{V}$. For any state $s_t$, information of nodes in $\mathcal{V}$, as encoded in their representations, is vital in determining the compatibility of two nodes for next edge creation.
To capture this insight, we propose a novel \textit{continuous latent action space},
whereby an action is mapped to the creation of an edge between two nodes with feature representation most similar to the action vector (\Cref{sec:action-selection}). 
In contrast to previous RL approaches to modeling graph structured data that define a discrete action space \citep{graph2seq2018,DasDhuZahVil18,YouLiuYinPanLes18}, our continuous latent action is independent of the size of graph, thereby facilitating scalable learning.

{\bf Transition Dynamics.} 
The transition function $P(s_{t+1}|s_t, a_t)$ is defined such that an action mapped to edge $(v_i, v_j)$ chosen at a state $s_t = (\mathcal{V}$, $\mathcal{E}_t)$ produces a next state $s_{t+1} = (\mathcal{V}$, $\mathcal{E}_t \cup (v_i, v_j))$. All edges are allowed for selection except when transition is a self-loop---i.e., both action components map to same node---which is rejected with no change in state. 

{\bf Reward} $\Rcal$. The GF-MDP perspective gives a concrete instantiation of the key component of our model -- an underlying (latent) optimization objective $\mathcal{F}_{\rm opt}$ in \eqref{eq:obj-minmax}:
$\Fcal_{\rm opt}$ is exactly the expected return $\mathbb{E}_{\pi_{\phi},P}[\sum_{t=0}^T \Rcal(s_t)]$ for executing a policy $\pi_{\phi}(a_t|s_t)$ with transition function $P(s_{t+1}|s_t,a_t)$ under a latent reward function $\Rcal$ evaluated at every state.
In contrast to existing RL frameworks for modeling graph structured data \citep{Xi2018,YouLiuYinPanLes18}, which use specific forms on the reward function, we propose to learn $\Rcal$ directly from the observed graph. 

Optimization over $\Fcal$ in \eqref{eq:obj-minmax} is the search for a reward function $\Rcal$ that assigns greater value to the observed graph than all other generated graphs, which reflects the assumption of optimality of the observed graph.
The optimization over $\Pi$ in \eqref{eq:obj-minmax} is an optimization over $\pi_{\phi}$ to maximize the expected return of $\Rcal$ over the formation process, which serves to construct a graph with minimal difference in structural properties, as measured by the reward, from the observed graph.
\vspace{-0.1cm}

\begin{figure*}
\centering
  
  \resizebox{1\textwidth}{!}{
  \begin{tabular}{c|c}
    \includegraphics[width=0.55\textwidth,height=0.3\textwidth]{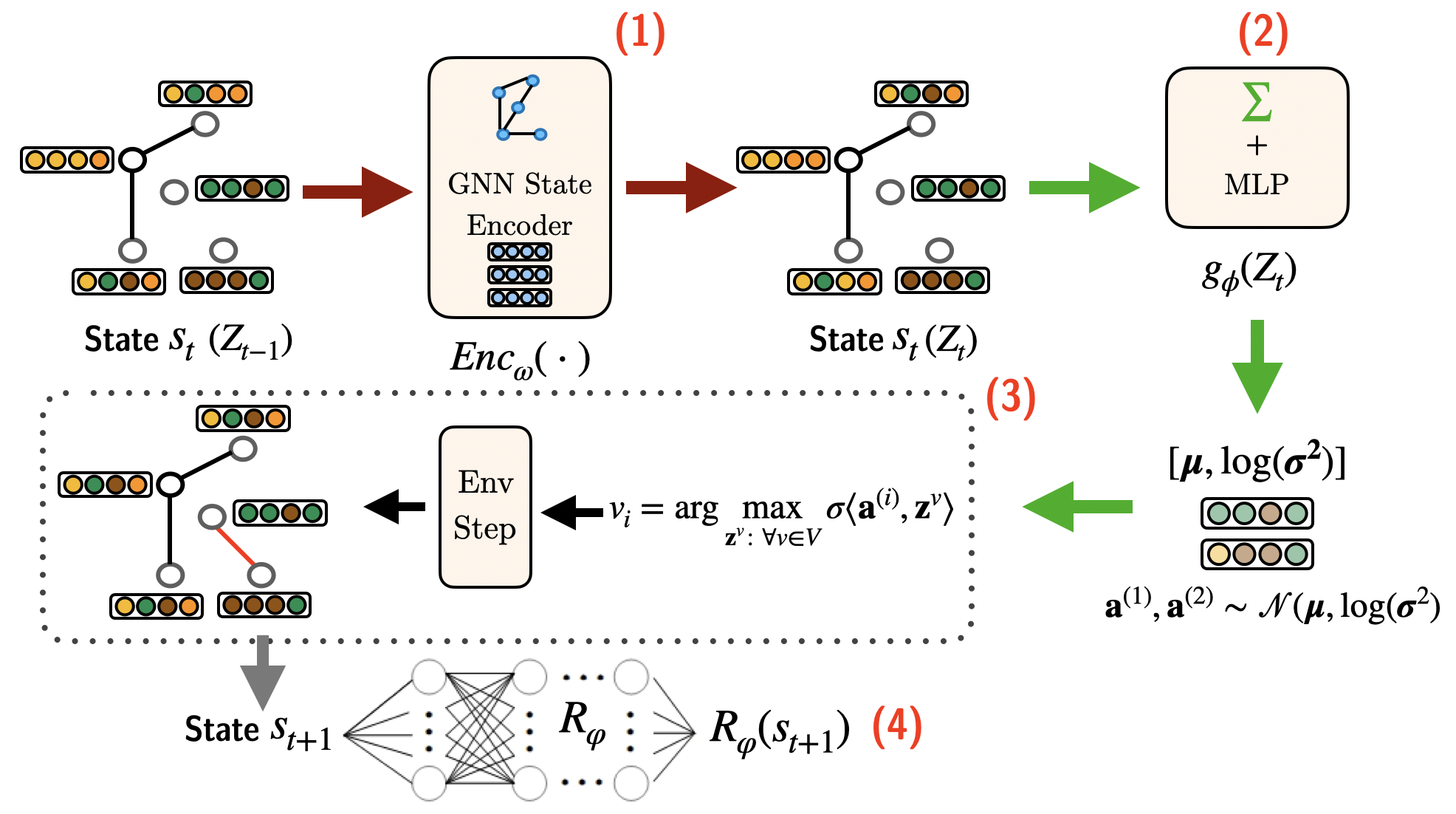} &
    \includegraphics[width=0.45\textwidth,height=0.25\textwidth]{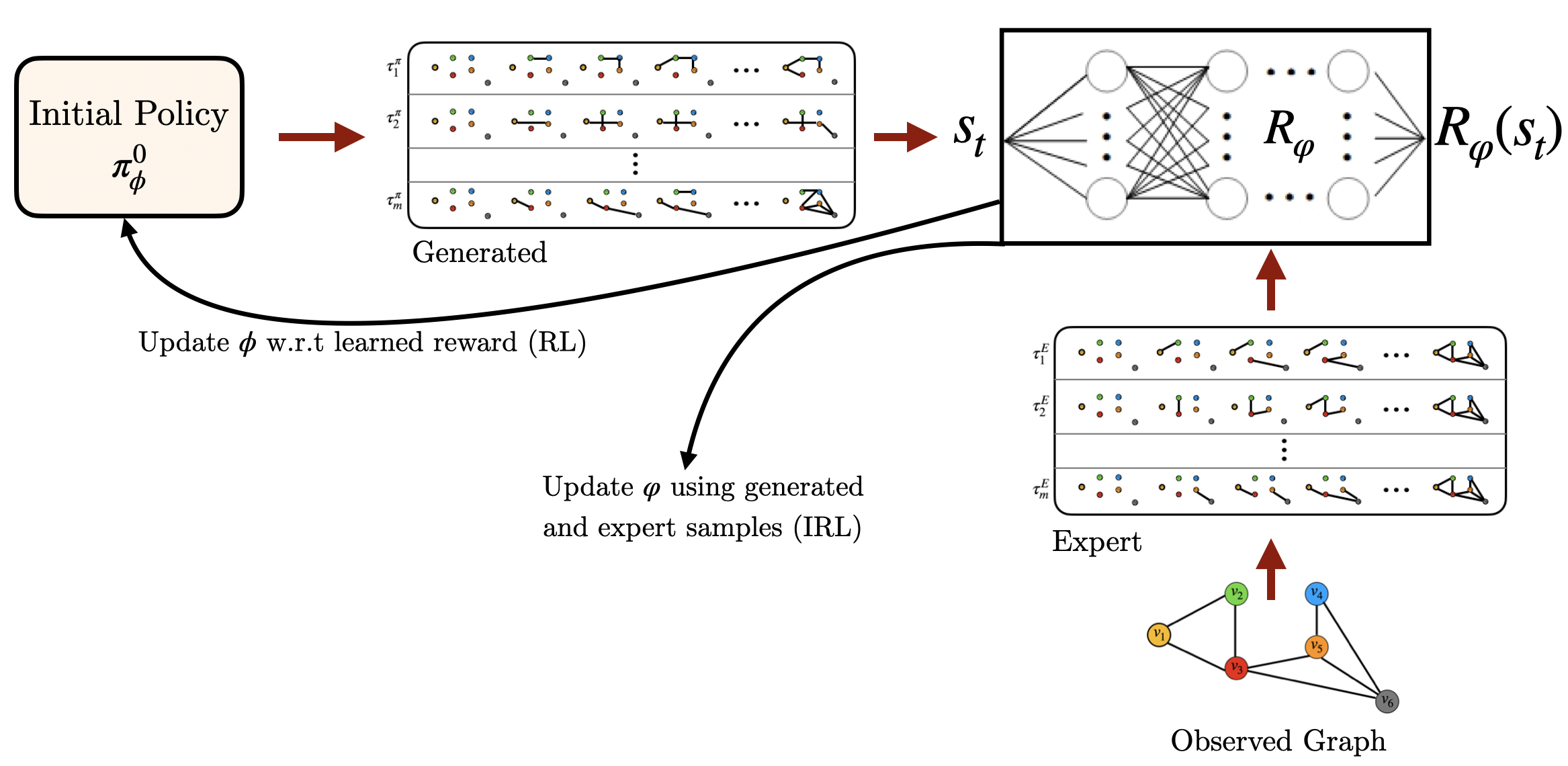} \\
  (a) GraphOpt Neural Policy Architecture & (b) GraphOpt Learning Loop
  \end{tabular}}
  \caption{Overview of GraphOpt Framework. (a) A GNN encoder maps a graph state $s_t$ into a representation $Z_t$ (1), which is aggregated and passed through an MLP (2), and interpreted as the mean and standard deviation of a Gaussian policy. 
  A latent continuous action $(\abf^{(1)},\abf^{(2)})$ is sampled and mapped to two nodes with most similar embeddings (3). States are evaluated by reward function $R_{\varphi}$ (4).
  (b) GraphOpt interleaves policy improvement using the current reward function and reward updates using generated and expert trajectories.}
\label{fig:architecture}
\end{figure*}

\vspace{-0.2cm}
\subsection{GraphOpt's Neural Policy Architecture}

As GraphOpt operates in a graph-structured environment, we build a graph neural network (GNN) based structured policy network to effectively utilize the structural information.
At time step $t$, GNN encodes graph state $s_t$ into low dimensional representation
for the policy to compute a corresponding action $a_t$.
We first describe the action selection procedure followed by the state encoder architecture. Figure~\ref{fig:architecture} (a) provides an overview of the policy network.

\subsubsection{Stochastic Action Selection} 
\label{sec:action-selection}

We design a stochastic policy that takes as input state $s_t$ and outputs a link formation action $a_t$.
As outlined in GF-MDP, we introduce a novel \textit{continuous latent action space} which induces 
action over node representations learned from data.
Specifically, action $a_t$ is a 2-tuple $(\abf^{(1)}, \abf^{(2)})$ whose components $\abf^{(i)} \in \Rbb^d$ are mapped to the node representations so as to select two nodes to construct an edge.
Let $v \in \mathcal{V}$ denote a node and let $\mathbf{z}^v \in \mathbb{R}^d$ denote its 
embedding 
(learned 
in \Cref{sec:state-encoder}).
Under a Gaussian policy $\pi_{\phi}$,
the next action is computed as follows:
\begin{equation}
    \begin{split}
    \label{eq:action-tuple}
        [\boldsymbol{\mu}, \log(\boldsymbol{\sigma^2})] = \pi(s_t) = g_\phi(\text{Enc}_{\omega}(s_t)) \\
        \abf^{(1)}, \abf^{(2)} \sim \mathcal{N}(\boldsymbol{\mu}, \log(\boldsymbol{\sigma}^2))
    \end{split}
\end{equation}
where $g_\phi$ is a two layer MLP with the policy parameters $\phi$.
$\text{Enc}_{\omega}(\cdot)$ is a state encoder 
that computes low-dimensional representations of graph states.
For an effective encoding of state information, we employ a GNN architecture with parameters $\omega$ (\Cref{sec:state-encoder}). 
Then we select two nodes to construct an edge using a similarity criterion: 
\vspace{-0.1cm}
\begin{equation}
    \label{eq:edge-selection}
    v_i = \argmax_{\zbf^v \colon \forall v \in V} \sigma \langle \abf^{(i)}, \zbf^v \rangle \quad \text{for $i = 1,2$}
\end{equation}
where $\langle \cdot, \cdot \rangle$ is a dot product and $\sigma$ is the sigmoid function. 
As the mapping from continuous $\abf^{(i)}$'s to node indices is external to policy network, GraphOpt is fully differentiable.

\vspace{-0.2cm}
\subsubsection{Structured State Encoder}
\label{sec:state-encoder}

During the graph formation process, the present structure of the graph may be a crucial factor that determines a new edge creation. 
Structural information of graphs are often encoded into low-dimensional representations and input to the policy network \citep{Wang18}. 
To achieve this, each state $s_t$ is represented by a node embedding matrix $\mathbf{Z}_t \in \mathbb{R}^{n \times d}$ (where $n = |\Vcal|$), computed using a GNN~\cite{Sca09}
via a $p$-step message propagation architecture.
At initial state $s_0$ when $\Ecal = \emptyset$, $\mathbf{Z}_0$ is initialized with node features.
After adding an edge at time $t$, we perform $p$ iterations of message passing across the node set to obtain $\mathbf{Z}_{t+1}$. For each iteration $p$, we update representation of each node as per the following equations:

Aggregate messages from the neighborhood of $v$: 
\begin{center}
$\mathbf{m}^p_v \leftarrow \textsl{AGG}(M(\mathbf{H}^{p-1}_{u})),~~\forall u \colon \mathbf{A}_t(u,v) = 1$ 
\end{center}

and then compute representation update for $v$ using:

\begin{center}
$\mathbf{H}^p_v \leftarrow \textsl{U}(\mathbf{H}^{p-1}_{v},\mathbf{m}^p_v)$
\end{center}

where $\mathbf{A}_t$ is the adjacency matrix. We use max pooling as $AGG$ aggregation function due to its better empirical performance. Both the message function $M$ and the update function $U$ are MLP. At the end of each training episode, we reset $\mathbf{Z}_{t}$ to initial state when resetting the environment.

\section{Maximum Entropy Learning Procedure}
\label{sec:maxent}

GraphOpt contains three modules: a graph construction policy $\pi$, a latent reward function $\mathcal{R}$, and a state encoder network\footnote{The state encoder network is trained by back-propagating the policy gradients to GNN parameters in an end-to-end manner.}. 
As the policy and latent reward are learned simultaneously in a graph structured environment, we require both stability and efficiency, which are difficult to satisfy simultaneously by off-policy methods such as DDPG\\~\cite{lillicrap2015continuous} and on-policy methods such as PPO~\cite{SchWolDhaRadKli17}.
To this end, we adopt Soft-Actor-Critic (SAC)
~\cite{haarnoja2018soft}, a maximum entropy variant of the actor-critic framework \citep{konda2000actor}, and combine it with maximum entropy based Inverse Optimal Control (IOC) objective~\cite{finn2016guided}.
We build a unified training pipeline that optimizes  following objectives:

(a) {\bf Soft Q-function.} SAC trains a function $Q_\theta(s_t, a_t)$ on off-policy experiences by minimizing the Bellman residual \[J_Q(\theta) = \mathbb{E}_{(s_t,a_t) \sim B} \bigl[ \bigl( Q_\theta(s_t,a_t) - \hat{Q}(s_t,a_t) \bigr)^2/2 \bigr]\]
where $\hat{Q}(s_t,a_t) = r(s_t, a_t) + \gamma \mathbb{E}_{s_{t+1}\sim p}[V_{\bar{\psi}}(s_{t+1})]$.
Value function $V_{\bar{\psi}}$ is implicitly defined by parameters of $Q_{\theta}(s,a)$\citep[Equation~3]{haarnoja2018soft}.

\begin{algorithm}[t!]
  \caption{GraphOpt Algorithm}
  \label{alg:GraphOpt}
  \begin{algorithmic}[1]
    \Procedure{GraphOpt}{}
     \State {\bf Input:} Empty trajectories list $\Tcal_{gen}$, replay buffer $B$
     \State node representation matrix $\mathbf{Z}_0$, parameters 
     \State $\psi, \phi, \theta, \omega, \varphi$. 
    \For{each epoch}
    \State Reset adj. matrix $\mathbf{A}_0 = \mathbf{0}$
    \State \textit{\# Using state encoder,}
    \State Reset state to $s_0 = \text{Enc}_{\omega}(\mathbf{Z}_0, \mathbf{A}_0)$
    \State Initialize trajectory $\tau = \lbrace s_0 \rbrace$
    \For{each environment step}
    \State $(v_1, v_2) \leftarrow a_t \sim \pi_\phi(a_t|s_t)$ using Eq~\ref{eq:action-tuple},~\ref{eq:edge-selection} 
    \State Update $\mathbf{A}_{t+1} \leftarrow \mathbf{A}_t[v_1,v_2]=1$.
    \State Update $s_{t+1} = \text{Enc}_{\omega}(s_t, A_{t+1})$
    \State Compute $r_t = R_{\varphi}(s_{t+1})$
    \State $B \leftarrow B \cup \{(s_t, a_t, r_t, s_{t+1}, A_{t+1})\}$
    \State Update trajectory $\tau \leftarrow \text{concat}(\tau , s_{t+1})$
    \State Train\_Policy ($B, \psi, \phi, \theta, \bar{\psi}, \omega$) \textit{\# Alg~\ref{alg:TrainPolicy}}
    \State \textbf{If} each edge in $G_t$ is repeated $k$ times or 
    \State max\_path\_length reached \textbf{then} 
    \State reset episode, store $\tau$ in $\Tcal_{gen}$, 
    \State and start new trajectory $\tau = \lbrace s_0 \rbrace$
    \EndFor
    \State Collect trajectories $\Tcal_{meas}$ (expert)
    \State Use $\Tcal_{meas}$ and $\Tcal_{gen}$ to update reward 
    \State estimator $R_\varphi$
    \EndFor
    \EndProcedure
  \end{algorithmic}
\end{algorithm}

(b) {\bf Policy Network.} The policy network $\pi_{\phi}(a_t | s_t)$ is trained using the following objective function: $$J_\pi(\phi) = \mathbb{E}_{s_t \sim B, \epsilon_t \sim \mathcal{N}}[\alpha\log\pi_\phi(a_t|s_t) - Q_\theta(s_t, a_t)].$$ Following~\cite{haarnoja2018soft}, we also use a reparameterization trick with a neural network transformation as: $a_t = f_\phi(\epsilon_t;s_t)$ that results in low variance estimator.

(c) {\bf Reward function.} $R_{\varphi}(s_t)$ is learned using the inverse optimal control objective:
$J_R(\varphi) := - \frac{1}{N} \sum_{\tau_i \in \Tcal_{\text{meas}}} \Rcal_{\varphi}(\tau_i) + \log \bigl( \frac{1}{M} \sum_{\tau_j \in \Tcal_{\text{gen}}} z_j \exp(\Rcal_{\varphi}(\tau_j)) \bigr)$
where $\mathcal{T}_{gen}$ is the set of link formation trajectories obtained from the learned policy and $\mathcal{T}_{meas}$ is the set of link formation trajectories obtained from the observed graph.
These measured trajectories are collected by accumulating edges from different permutations over the ordering of edges in the original graph. 
All permutations can be considered ``expert'' trajectories as each starts from same initial state ($\Ecal_0 = \emptyset$) and contains only true edges seen in the observed graph.

\Cref{alg:GraphOpt} outlines the complete set of steps that are used for end-to-end training of GraphOpt. An epoch starts with initial state representation computed using state encoder when there is no edge between the nodes (line 4-7). 
For every step in the environment, the policy either creates a new edge or repeats an existing edge (line 8).
It receives a reward based on current reward function and the state of the environment is updated along with replay buffer and current trajectory information (line 9-13).
We train the policy network, Q-network, and state encoder after every few steps taken by the environment (line 15).
If all existing edges have repeated $k$ times, the episode ends (line 16-19), the environment is reset and new trajectory $\tau$ is initialized.
The reward network is trained after end of each epoch (line 21-23). \Cref{app:gradient-updates} details the gradient updates.
In contrast to generative adversarial approaches to \eqref{eq:obj-minmax} for imitation learning \citep{ho2016generative}, which converge to an uninformative discriminator, and in contrast to behavioral cloning \citep{torabi2018behavioral}, which does not provide an explanatory mechanism, maximum entropy IRL satisfies the key objective of our work by recovering a useful latent reward. Figure~\ref{fig:architecture} (b) provides an overview of our algorithm.
\vspace{-0.3cm}
\section{Experiments}
In this section, we aim to answer the following questions to evaluate the efficacy of our approach:

(i) Can GraphOpt discover a useful latent objective that is operationally equivalent to some underlying mechanism of a graph-structured domain, and thereby \textit{transferable} to an unseen graph in that domain?
The success and necessity of transfer is shown by how well the objective discovered on a source graph facilitates construction when optimised by a new policy on a target graph, in contrast to directly running (i.e., without fine-tuning) the trained policy  on the target.

(ii) How well does GraphOpt's construction policy, learned by optimizing the discovered objective, \textit{generalize} to downstream inference tasks?
Here, we turn to the classical problem of link prediction in graphs,
which requires the construction policy to generalize to  predict hidden links with high accuracy, and---more crucially---perform a link prediction task for which it was not explicitly trained.
We further stress test GraphOpt's generalization capacity by deploying a trained policy on unseen target graph environments of different size and characteristics, without fine-tuning.

(iii) Can GraphOpt serve as an effective graph constructor to synthesize new graphs that exhibit structural patterns similar to those in an observed graph?
We assess the performance of the learned construction policy by deploying it on the full set of training nodes and generating new graphs, analogous to standard practice in reinforcement learning~\cite{sutton2018reinforcement}.
GraphOpt's stochastic policy avoids copying the observed graph while preserving the statistical properties.

To the best of our knowledge, no single baseline can do all three tasks.
Hence we compare GraphOpt with task-specific baselines for (ii) and (iii) and follow standard procedure in the IRL literature to report performance for (i).

\textbf{Training.} 
All experiments begin by using the observed graph to learn the construction policy and latent reward function via \Cref{alg:GraphOpt}.  A key advantage of using SAC as the base RL algorithm is that it largely eliminates the need for per-task hyperparameter tuning.
To encourage creation of new edges during training, we terminate an episode when the number of repeated creations of \textit{each} existing edge reaches a threshold $k$, which signifies that the policy has lost the ability to explore further. 
We provide more details on other training configurations in \Cref{app:impl}.

\subsection{Discovering Transferable Latent Objective}

For this experiment, we use two citation graphs: Cora-ML as a source environment and Citeseer as a target environment. 
We train on Cora-ML to discover a latent reward function, which is then transfered to train a new policy network from scratch on CiteSeer. While training on Citeseer, the reward function remains fixed and is not further trained.
\vspace{-0.3cm}
\begin{figure}[ht!]
\small
\centering
\begin{tabular}{cc}
\includegraphics[width = 0.23\textwidth]{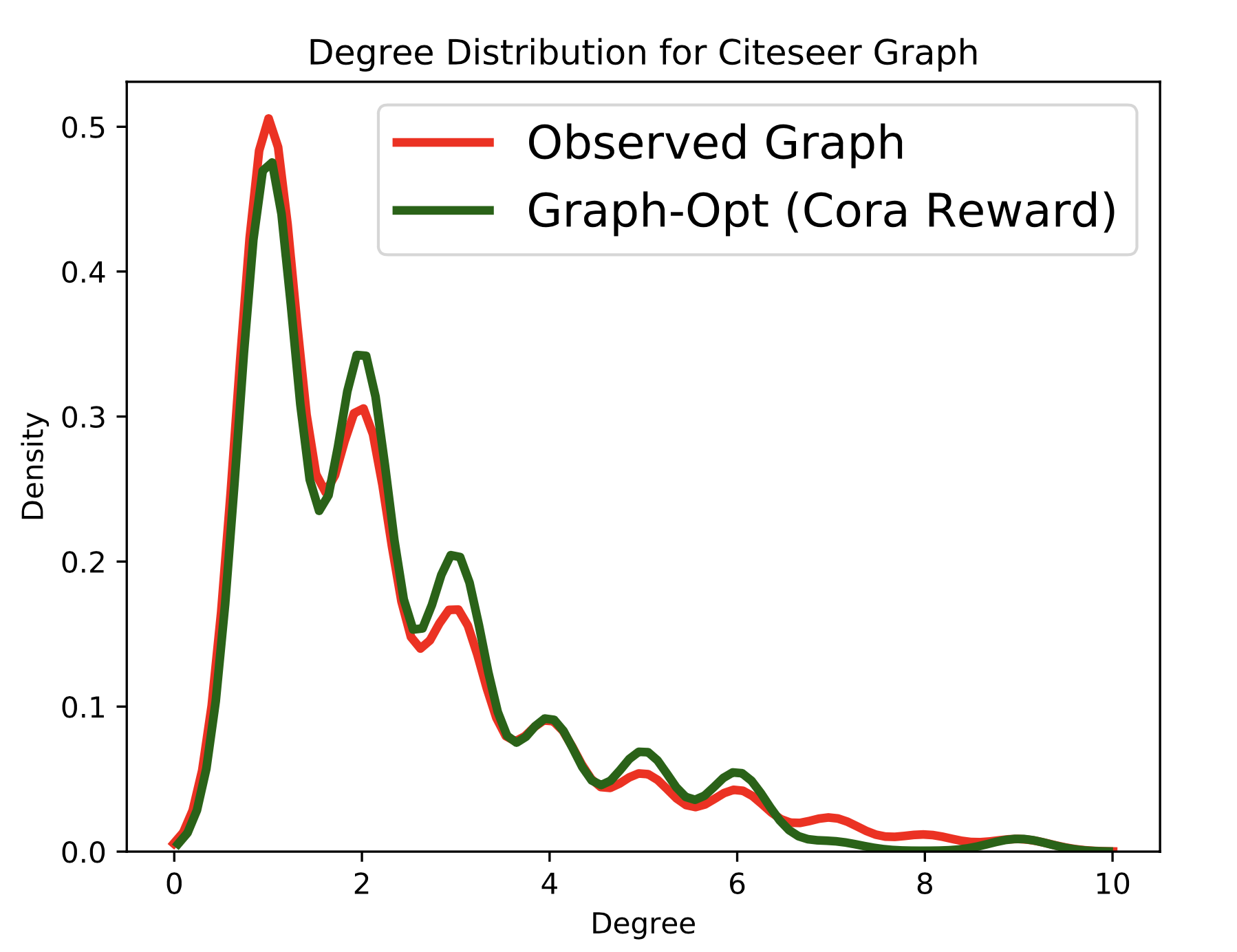}
& \includegraphics[width = 0.23\textwidth]{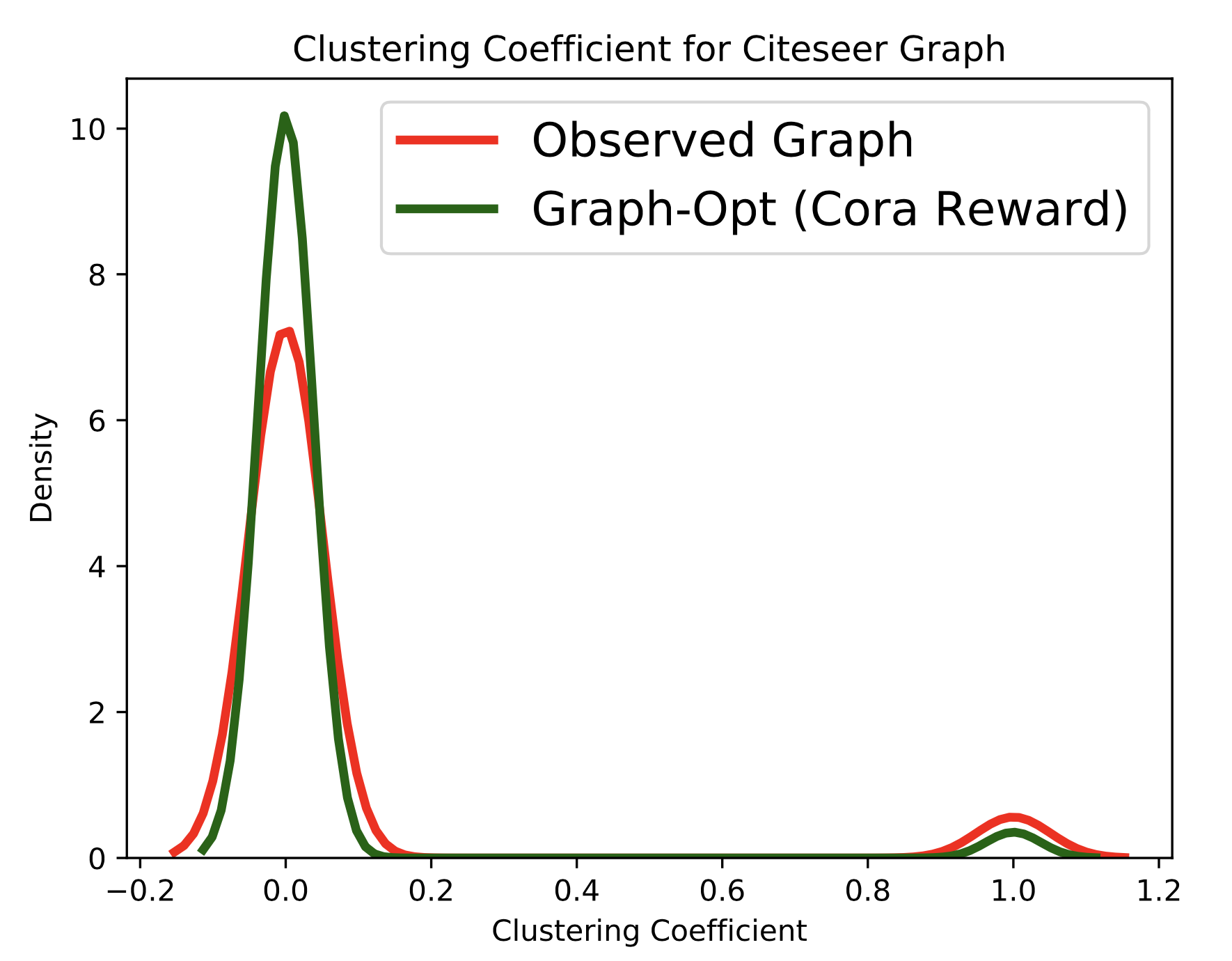}\\
(a) Degree distribution & (b) Clus Coeff distribution
\end{tabular}
\vspace{-2mm}
\caption{Degree and Clus. Coeff. distribution of graph constructed using the policy learned on CiteSeer, while optimizing the objective transferred from training on Cora-ML dataset.}
\label{fig:transfer_plots}
\vspace{-0.8cm}
\end{figure}

\begin{table}[ht!]
\captionsetup{font=footnotesize}

\centering
  \caption{Transfer Performance Comparison}
\resizebox{0.45\textwidth}{!}{
    \begin{tabular}{cccc}
    \toprule
          & \textbf{Triangle Count} & \textbf{Clustering Coeff.} & \textbf{Max Degree} \\
    \toprule
    Cora-ML (train) & 4890  & 0.241 & 168   \\
    CiteSeer (observed) &  3501    & 0.1414     & 99     \\
    \midrule
    CiteSeer (reward) &   $2847.66 \pm	57.13$    &    $0.098\pm	0.0010$   & $80.66 \pm	1.527$       \\
    CiteSeer (direct) &    $2234 \pm	58.96$   &   
$0.084\pm	0.004$    &   $70.66 \pm	2.081$\\
    \bottomrule
    \end{tabular}}
  \label{tab:transfer_stats}%
  
\end{table}

After training, we input CiteSeer's node set with an empty edge set to the model and run the evaluation policy to construct edges. We collect 3 graphs and report mean and standard deviation for graph based statistics representing network patterns of the generated graphs. 
Table~\ref{tab:transfer_stats} and Figure~\ref{fig:transfer_plots} demonstrates that optimising the transferred objective, GraphOpt learns an effective policy to construct edge topologies that results in similar network patterns as observed in Citeseer.
In contrast, the poor performance of the Cora-trained policy when directly deployed on CiteSeer without training with the transferred objective (``CiteSeer (direct)'' in \Cref{tab:transfer_stats} and \Cref{fig:cora-trained-policy}) shows that the discovered objective is important for transfer.
This experiment demonstrates GraphOpt's effectiveness in discovering a useful latent objective that serves as an explanation for the formation of observed network patterns across citation graphs, thereby lending itself to transfer across graphs within this domain.

\subsection{Policy Generalization to Prediction Task}

We first discuss the task of link prediction, which demonstrates GraphOpt's ability to learn a construction policy that generalizes to unseen task for which it was not explicitly trained. 
We then show the performance of the policy to generalize to new graphs of different sizes and characteristics.

\subsubsection{Link Prediction}

{\bf Setup.} 
We conduct link prediction experiments on a variety of graphs from both non-relational and relational domains\footnote{ Table~\ref{tab:pred_graph1} and~\ref{tab:pred_graph2} in \Cref{app:datasets} provide dataset details}. 
We compare our performance with both explicit baselines that employ a dedicated hand-designed link prediction objective and implicit models that generalize to the link prediction task without using explicit link prediction objectives during training; GraphOpt falls under the latter category. 
For all experiments, we follow the protocol in ~\cite{seal2018} by randomly removing $10\%$ of edges to form a held-out test set and randomly sampling the same number of nonexistent links to form negative test samples. 
Training is then performed on the remaining graph. 
For relational graphs, we include edge type as an extra feature in the message passing scheme in the state encoder.

After training, we provide the observed graph as initial state and run the policy to assess how well it can predict hidden edges. 
For non-relational baselines, we label each edge from the test set as 1 if the policy created it and 0 otherwise, and compare with true labels to report AUC (Area under curve) and AP (average precision). 
For relational baselines, for a test triple ($(e_s, r, e_o)$), we collect all the edges that are created for tuple $(e_s, r)$ and rank them in order of creation to report MRR and HITS@10---this signifies policy's preference to create one test edge over another. We also perform link prediction experiments to assess the usefulness of node representations learned by our state encoder.

{\bf Performance.} Tables~\ref{tab:linkpred_policy_nonr} and~\ref{tab:linkpred_policy_r} show that
GraphOpt's link prediction performance surpasses implicit baselines on most datasets.
It is highly competitive with NetGAN (non-relational),which uses a GAN based objective, 
and shows significant improvement over RL based Minerva (relational), which uses LSTM encoder for state representation and a fixed reward value of +1/-1 for each step. 
Superior prediction performance demonstrates that GraphOpt learns a model with strong generalization capacity.
GraphOpt's success in this aspect can be attributed to the combination of a stochastic policy that encourages exploration during training and the ability of the GNN to encode state representations that generalize to inference time.
As a tradeoff for its greater generality, GraphOpt does not have the luxury of domain-specific architectures or objectives; hence its competitive but often slightly worse performance against state-of-the-art dedicated link prediction baselines is not surprising. 
However, GraphOpt's comparable performance demonstrates its greater potential for domains where an objective function is not known \textit{a priori} and hand-designing an objective is difficult---this opens up exciting avenues of research for improvement.
Surprisingly, the embedding based prediction shows strong performance, often surpassing baselines on various datasets (last row in \Cref{tab:linkpred_policy_nonr} and \Cref{tab:linkpred_policy_r}).
This demonstrates that GraphOpt learns a representation network that can be independently leveraged to perform various downstream tasks.
As the encoder is trained in the same computational graph as the policy, through optimising the discovered latent objective,  this further supports the usefulness of the discovered objective.

\begin{table}[t!]
\captionsetup{font=footnotesize}
\centering
\caption{Link Prediction on non-relational data: (*) is used to signify better performer amongst GraphOpt and method with implicit objective. Bold numbers are best two performers overall.}
\resizebox{0.48\textwidth}{!}{
    \begin{tabular}{lrrrrrr}
        \toprule
          & \multicolumn{2}{c}{\textbf{Cora-ML}} & \multicolumn{2}{c}{\textbf{Political Blogs}} & \multicolumn{2}{c}{\textbf{E. Coli}} \\
          \cmidrule(lr){2-3}\cmidrule(lr){4-5}\cmidrule(lr){6-7}
          & \multicolumn{1}{c}{\textbf{AUC}} & \multicolumn{1}{c}{\textbf{AP}} & \multicolumn{1}{c}{\textbf{AUC}} & \multicolumn{1}{c}{\textbf{AP}} & \multicolumn{1}{c}{\textbf{AUC}} & \multicolumn{1}{c}{\textbf{AP}} \\
          \toprule
          
    VGAE  &    94.70   &  96.10      &    92.60   & 93.44      &   93.22    & 93.10 \\
    Node2Vec & 91.12      & 91.78      &    87.22   &  85.51     &  79.99     & 74.32 \\
    NetGAN &    94.20*   &    95.22*   & \textbf{95.51}*      & 90.00      &    93.17   & 94.50 \\
    SEAL  & \textbf{97.21}      &    \textbf{97.99}   & 95.32      &  \textbf{96.10}     &  \textbf{97.12}     & \textbf{97.50} \\
    \midrule
    GraphOpt-Policy &  93.50    &  94.87     &  92.21     &     92.33*  &    94.43*   & 95* \\
    GraphOpt-Embed &    \textbf{96.21}   &  \textbf{96.66}     &  \textbf{95.50}     &  \textbf{95.32}     &     \textbf{97.20}  &  \textbf{97.44}\\
    \bottomrule
    \end{tabular}}
\label{tab:linkpred_policy_nonr}
\end{table}

\begin{table}[t!]
  \centering
  \caption{Link Prediction performance on relational data: (*) is used to signify better performer amongst GraphOpt and RL method with +1/-1 reward. Bold numbers indicate best two performers overall.}
  \resizebox{0.48\textwidth}{!}{
    \begin{tabular}{lrrrrrr}
    \toprule
          & \multicolumn{2}{c}{\textbf{Kinship}} & \multicolumn{2}{c}{\textbf{FB15K-237}} & \multicolumn{2}{c}{\textbf{WN18RR}} \\
          \cmidrule(lr){2-3}\cmidrule(lr){4-5}\cmidrule(lr){6-7}
          & \multicolumn{1}{c}{\textbf{MRR}} &  \multicolumn{1}{c}{\textbf{H@10}} & \multicolumn{1}{c}{\textbf{MRR}} &  \multicolumn{1}{c}{\textbf{H@10}} & \multicolumn{1}{c}{\textbf{MRR}} & \multicolumn{1}{c}{\textbf{H@10}} \\
    \toprule
    ConvE &     \textbf{87.1}     &     \textbf{98.1}  &     \textbf{43.5}  & \textbf{62.2}      & \textbf{44.9}      & \textbf{54} \\
    NeuralLP &    61.9     & 91.2      & 22.7     & 34.8      & 46.3  & 65.7 \\
    Reward Shaping &   \textbf{87.8}     &   \textbf{98.2}    &    \textbf{40.7}   &    56.4   &   47.2     &  54.2\\
    Minerva &   72.0   &    92.4*   &    29.3   &     45.6  &   44.8* & 51.3 \\
    \midrule
    GraphOpt-Policy &  82.2*    &   92.33    &   33.12*    &  53.22*     &   44.2     & 53.6* \\
    GraphOpt-Embed &  84 & 96.12 &	39.66 &	\textbf{58.51} &	\textbf{47.3} &	\textbf{58.43}  \\
    \bottomrule
    \end{tabular}}
    \vspace{-0.5cm}
  \label{tab:linkpred_policy_r}
\end{table}

\begin{table}[t!]
  \centering
  \caption{Generalization Performance Comparison}
  \resizebox{0.48\textwidth}{!}{
    \begin{tabular}{cccc}
    \toprule
          & \textbf{Triangle Count} & \textbf{Clustering Coeff.} & \textbf{Max Degree} \\
    \toprule
    BA-200 (train) & 780   & 0.12  & 43    \\
    BA-1000 (observed) & 1632   & 0.0407  & 115    \\
    \midrule
    BA-1000 (eval) & $1470.66 \pm	25.71$    &   $0.036\pm	0.0044$    & $119.33	\pm 2.081$   \\
    \toprule
    Cora (train) & 4890  & 0.241 & 168   \\
    CiteSeer (observed)  &  3501    & 0.1414     & 99     \\
    \midrule
    Citeseer (eval) &    $2234 \pm	58.96$   &   
$0.084\pm	0.004$    &   $70.66 \pm	2.081$    \\
    \bottomrule
     \end{tabular}}
  \label{tab:generalize_app}%
\end{table}%

\subsubsection{Evaluating Policy performance on Unseen Environments}

\begin{figure}[t!]
\centering
\begin{subfigure}[t]{0.5\linewidth}
    \centering
    \includegraphics[width =1.0\textwidth]{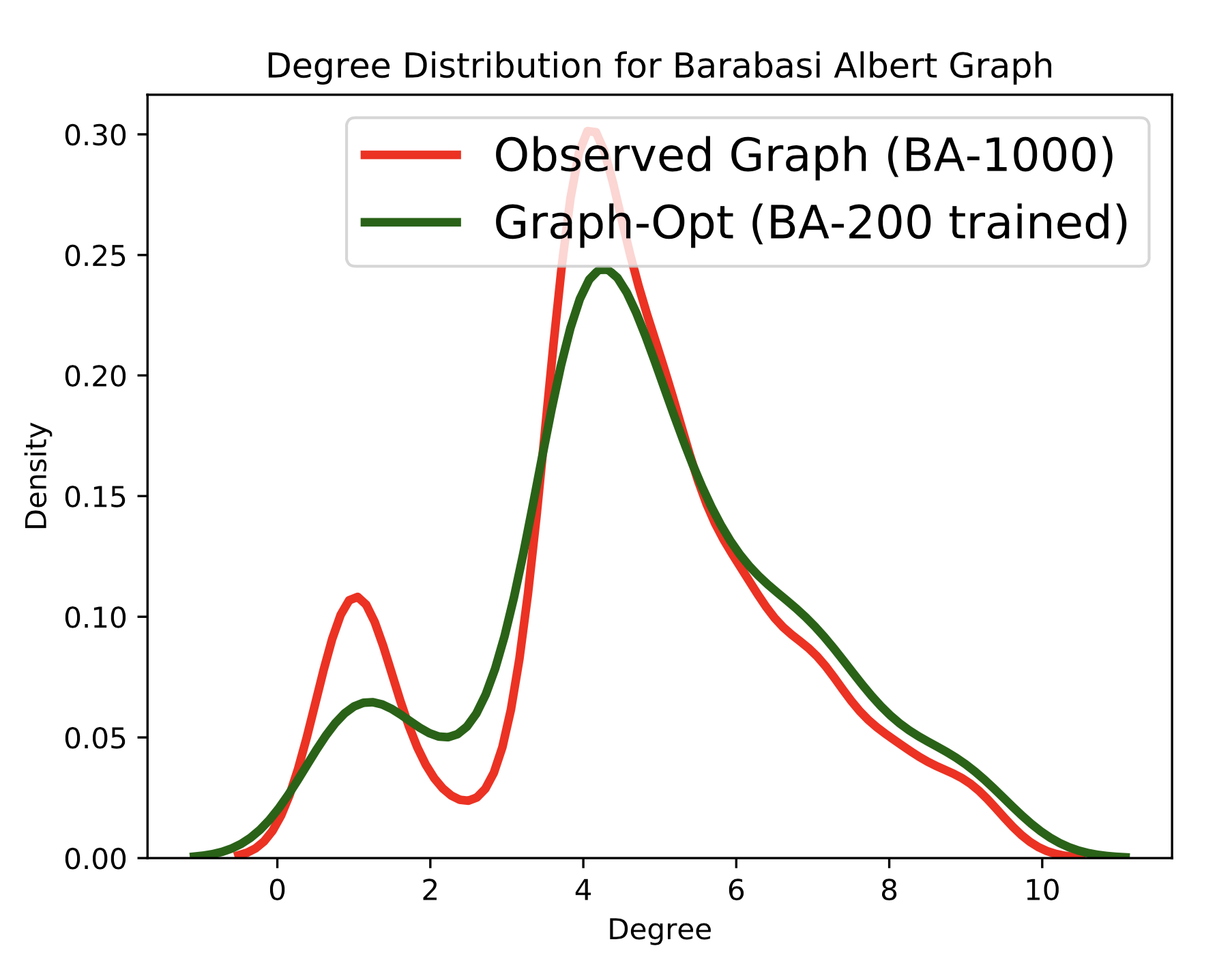}
    \caption{BA-200 trained policy}
    \label{fig:ba-200-trained}
\end{subfigure}%
\hfill
\begin{subfigure}[t]{0.5\linewidth}
    \centering
    \includegraphics[width = 1.0\textwidth]{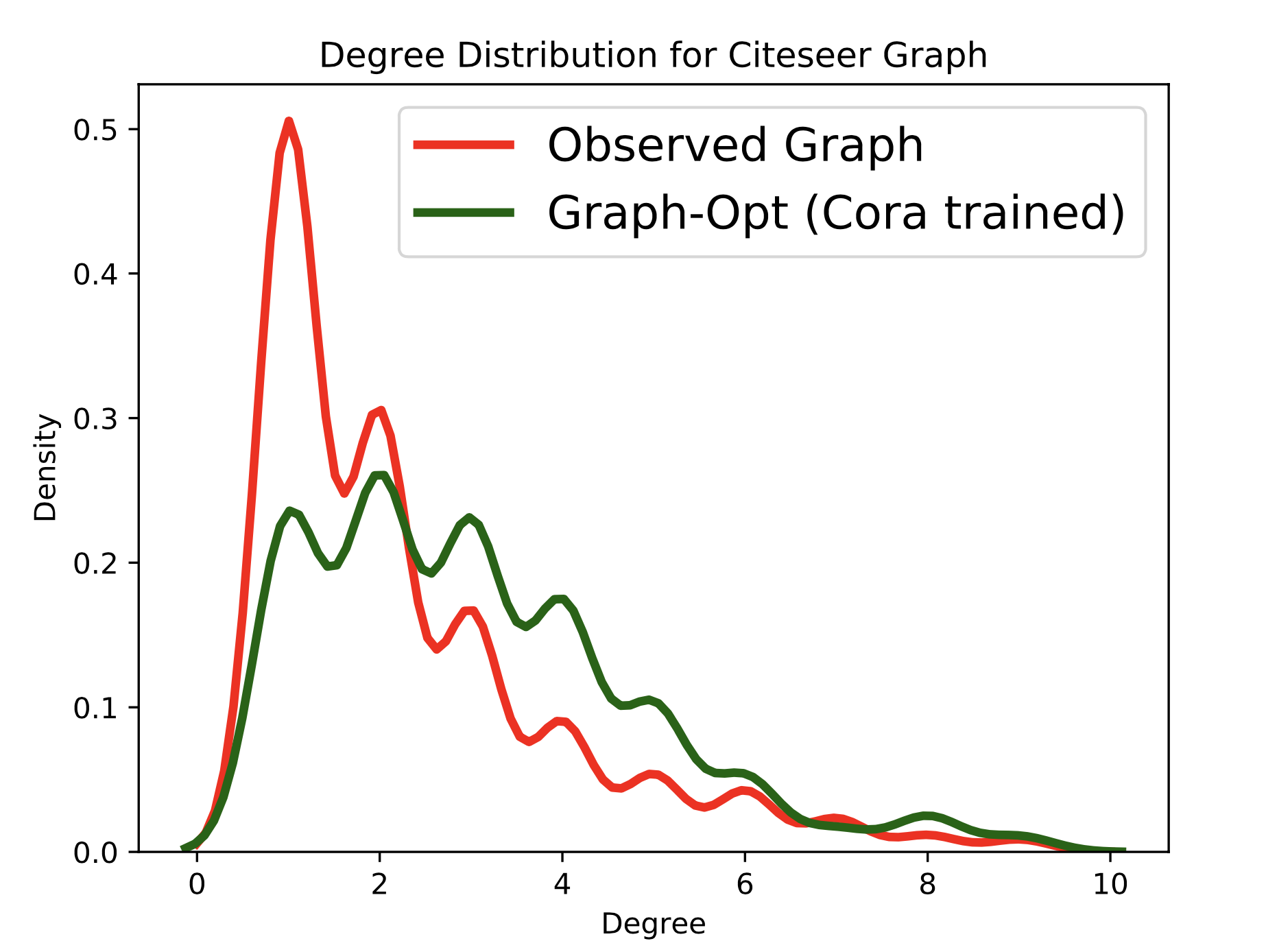}
    \caption{Cora trained policy}
    \label{fig:cora-trained-policy}
\end{subfigure}
\caption{Policy Transfer across different size (Barabasi-Albert graph) and different graph (Cora-ML$\rightarrow$Citeseer).}
\label{fig:generalize_plot}
\end{figure}

We now focus on the performance of the construction policy when trained on a source graph and deployed on an unseen target graph without further training\footnote{For these experiments, we are only interested in evaluating the learned policy on a target environment; hence the learned reward is not used on the target graph.}. 
Specifically, we investigate two aspects of this direct policy transfer:

\textit{(i) Transfer from small to large graph from same distribution.}
We train GraphOpt on a source BA graph of 200 nodes.
Then we input the node set of a target BA graph with 1000 nodes to the learned policy and run it without further training to generate 3 graphs. 
\Cref{fig:ba-200-trained} and \Cref{tab:generalize_app} (top 3 rows) demonstrate that generated graphs exhibit similar properties to the BA graph of 1000 nodes. 
This suggests that the learned construction policy can be used to generate synthetic graphs of larger size than the training graph, while preserving the underlying structural properties.
\textit{(ii) Direct transfer from source to target graph in the same domain.}
We return to the earlier experiment on citation graphs, but this time transferring the trained policy from Cora-ML to run on the CiteSeer node set, without using the learned objective.
As expected, \Cref{tab:generalize_app} (bottom 3 rows) and \Cref{fig:cora-trained-policy} show that the policy mostly fails to construct similar patterns as observed in the original Citeseer, but it does approximate some patterns (e.g. max degree) well. 
This supports the necessity of using GraphOpt's discovered reward for transfer and suggests room for further investigation.

\begin{table*}[ht!]
  \centering
  \caption{Percent deviation of graph statistics for generated graph from observed one (lower is better). First row displays the actual statistics of the observed graph. Results for more graphs and more metrics for generated graphs are available in \Cref{app:generative-capability}.}
  \resizebox{1\textwidth}{!}{
    \begin{tabular}{cccccccccc}
    \toprule
          & \multicolumn{3}{c}{\textbf{Barabasi Albert}} & \multicolumn{3}{c}{\textbf{Political Blogs}} & \multicolumn{3}{c}{\textbf{CORA-ML}} \\
          \cmidrule(lr){2-4}\cmidrule(lr){5-7}\cmidrule(lr){8-10}
    \textbf{Model} & \textbf{Triangle Cnt.} & \textbf{Clust. Coeff.} & \textbf{Max Degree} & \textbf{Triangle Cnt.} & \textbf{Clust. Coeff.} & \textbf{Max Degree} & \textbf{Traingle Cnt.} & \textbf{Clust. Coeff.} & \textbf{Max Degree} \\
    \toprule
    Observed Graph & 504 & 0.1471 & 33 & 303129  & 0.319 & 351 & 4890  & 0.2406 & 168 \\
    \midrule
    DC-SBM & $46.56 \pm 6.58$  & $59.44 \pm 7.11$ & $28.29 \pm 7.63$ & $52.78 \pm 9.15$  &$91.73 \pm 1.18$ & $40.86 \pm 1.89$ & $71.17 \pm 1.53$  & $68.25 \pm 20.16$ & $6.94 \pm 5.40$ \\
    BTER  & $48.02 \pm 9.11$  & $33.20 \pm 1.28$& $33.33 \pm 0$& $45.47 \pm 7.25$  & $54.17 \pm 13.57$ & $43.87 \pm 0.75$ & $40.06 \pm 1.17$  & $81.66 \pm 1.74$ & $16.47 \pm 14.49$\\
    VGAE  & $70.89 \pm	8.95$ & $94.40 \pm 0.81$ & $8.08 \pm 1.75$  & $98.56 \pm 0.44$ & $99.32 \pm 0.55$ & $44.06 \pm 0.92$  & 
$99.56 \pm 0.24$ & $93.10 \pm 2.11$ & $94.44 \pm 1.82$ \\
    NetGAN & $31.68 \pm	6.28$& $40.69 \pm 4.27$& $\mathbf{4.04 \pm 1.74}$  & $44.28 \pm 8.27$ & $37.55 \pm 7.2$ & $\mathbf{38.75 \pm 3.70}$  & $64.19 \pm 2.15$ & $41.12 \pm 18.82$ & $4.17 \pm 2.38$\\
    \midrule
    {\bf GraphOpt} & $\mathbf{6.28 \pm	4.05}$ &$\mathbf{25.52 \pm 8.25}$ & $\mathbf{5.05 \pm 4.63}$ & 
$\mathbf{34.73 \pm 3.79}$  &$\mathbf{20.34 \pm 9.1}$ & 
$\mathbf{36.85 \pm 2.71}$& 
$\mathbf{19.46 \pm 1.01}$ & 
$\mathbf{14.63 \pm 5.78}$ & $\mathbf{2.58 \pm 1.24}$\\
    \bottomrule
    \end{tabular}}%
  \label{tab:gen_result}%
\end{table*}
\subsection{Synthesizing Graphs via Learned Generative Mechanism}

{\bf Setup.} 
We evaluate GraphOpt's ability to synthesize new graphs after learning to construct from an observed graph.
We use both synthetic and real world graphs that span different domains, characteristics and sizes (Table~\ref{tab:datastat_gen} in \Cref{app:datasets}).
All graphs are undirected. For training, we use the complete observed graph. 
For evaluation, we provide the node set of the input graph and empty edge set and run the trained policy to construct edges. 
GraphOpt learns from a single large graph and hence we compare with strong baselines with similar setting (details in \Cref{app:baselines}).
For all methods, we generate 3 graphs for evaluation and report mean and standard deviation of percentage error of graph based statistics between observed and generated graphs (Table~\ref{tab:gen_result}). 
For GraphOpt, we run the evaluation policy up to either the original termination condition or a multiple of actual number of edges in observed graph, whichever is earlier. 
We follow reported stopping criteria for baselines.
\begin{figure}[t!]
    \centering    
    \begin{subfigure}[t]{0.5\linewidth}
        \centering
        \includegraphics[width=1.0\textwidth]{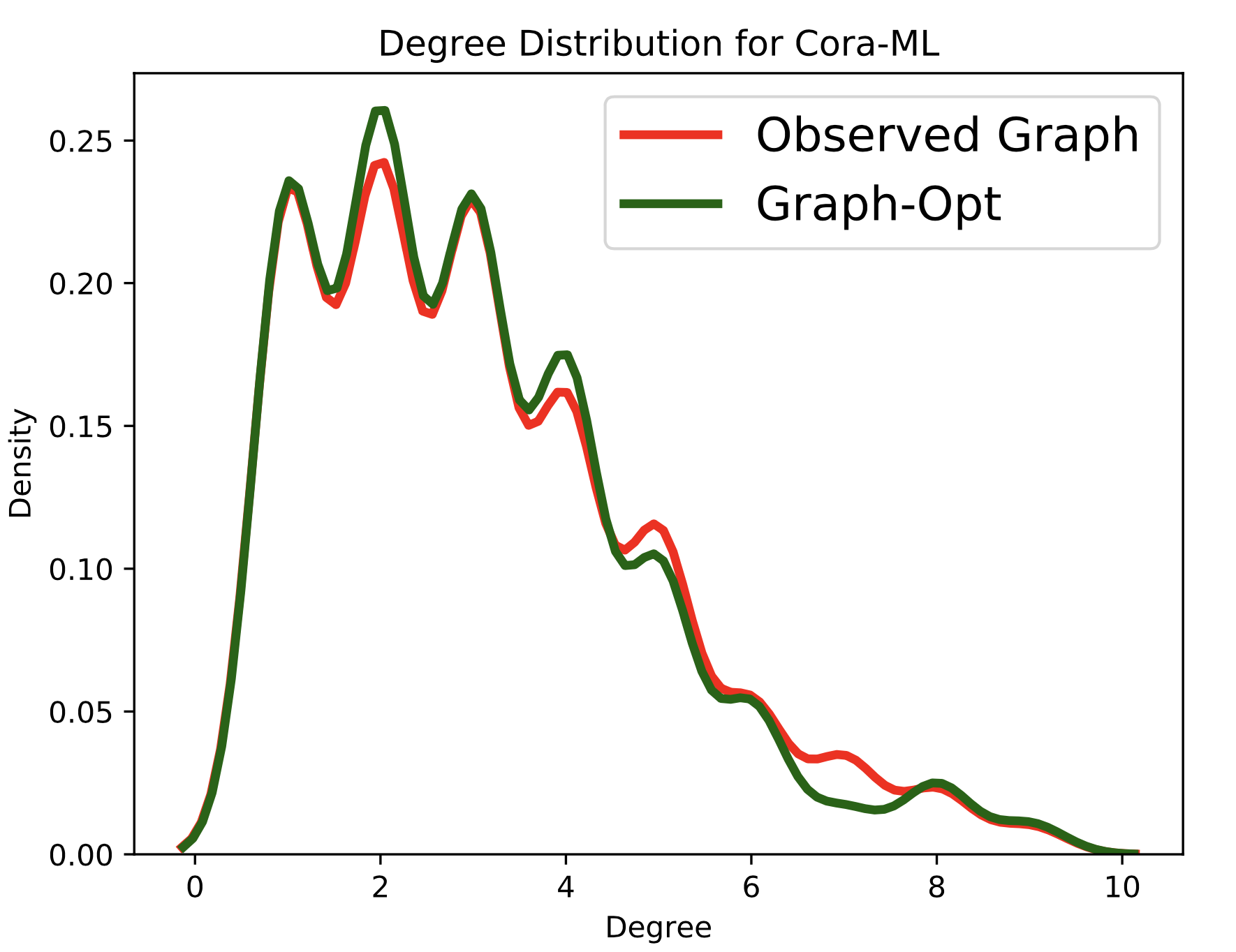}
        \caption{Degree distribution}
        \label{fig:cora-degree-distribution}
    \end{subfigure}%
    \hfill
    \begin{subfigure}[t]{0.5\linewidth}
    \centering
        \includegraphics[width=1.0\textwidth]{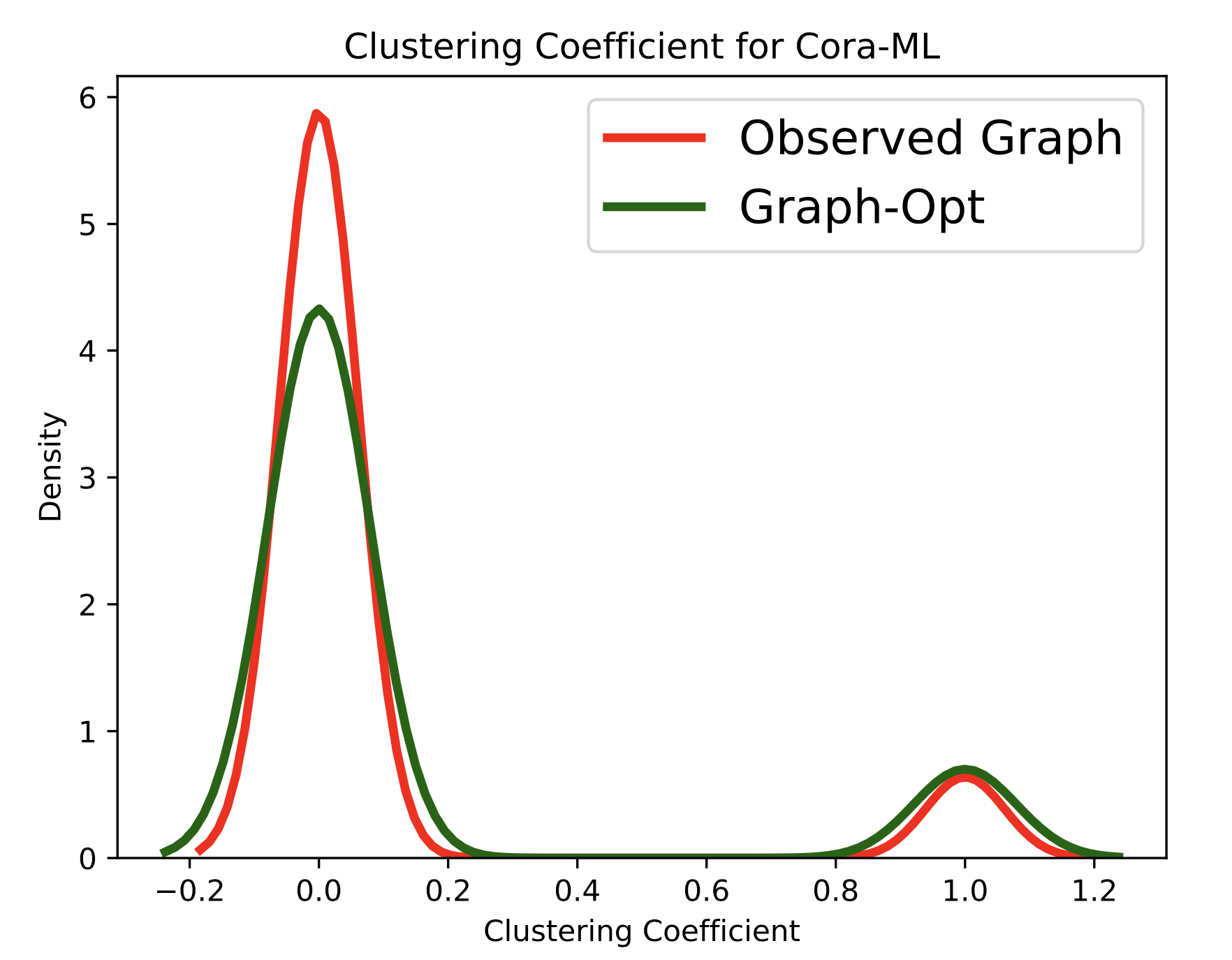}
        \caption{Clus. Coeff. distribution}
        \label{fig:cora-clustering}
    \end{subfigure}
    
    \begin{subfigure}[t]{0.49\linewidth}
    \centering
        \includegraphics[width=1.0\textwidth]{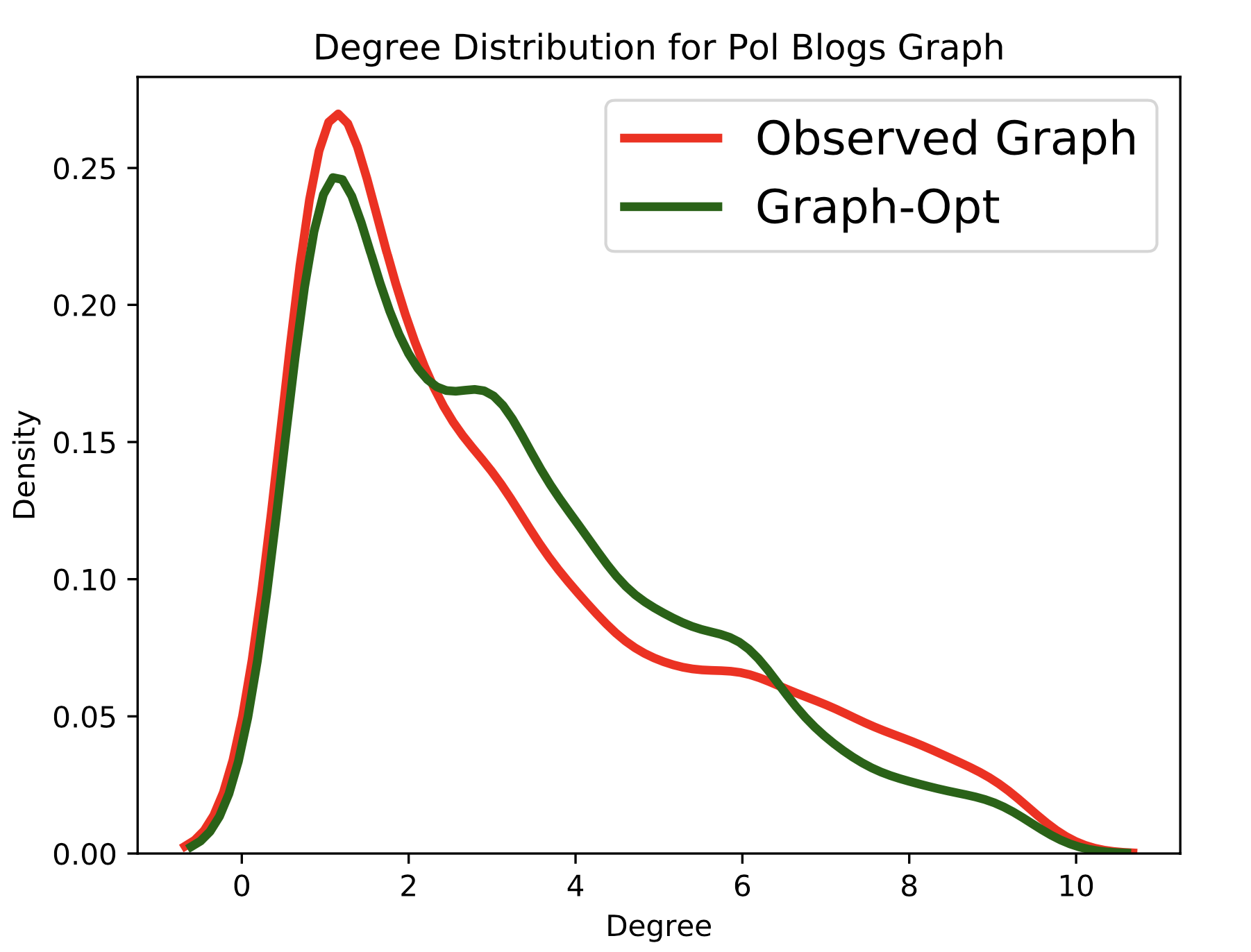}
        \caption{Degree distribution}
        \label{fig:pol-degree-distribution}
    \end{subfigure}
    \hfill
    \begin{subfigure}[t]{0.49\linewidth}
    \centering
        \includegraphics[width=1.0\textwidth]{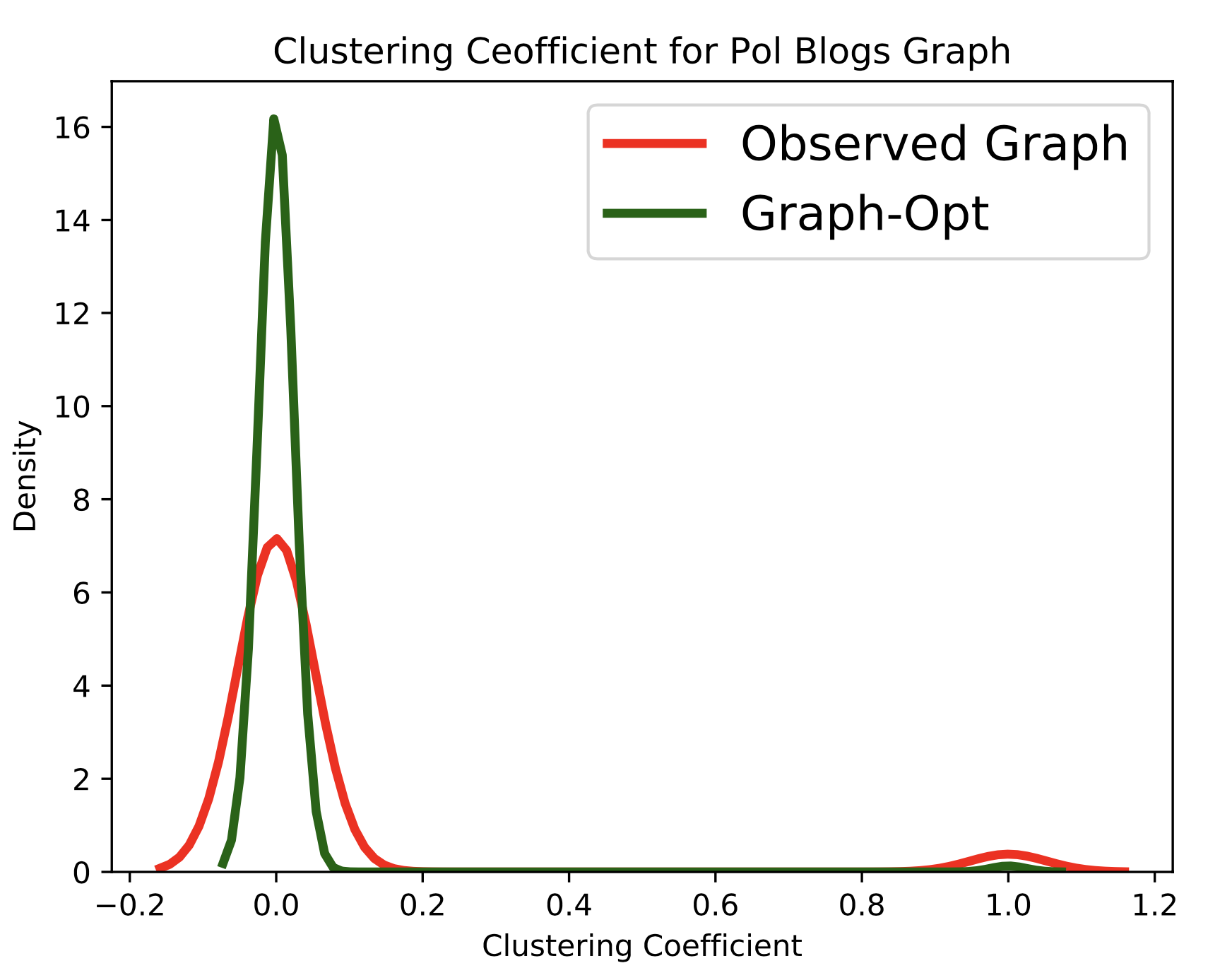}
        \caption{Clus. Coeff. distribution}
        \label{fig:pol-clustering}
    \end{subfigure}
    \caption{{\bf (a-b}) Original vs. GraphOpt: Cora-ML ({\bf c-d}) Original vs. GraphOpt: Pol.Blogs}
    \label{fig:coar_pol}
    \vspace{-0.5cm}
\end{figure}

{\bf Performance.} Table~\ref{tab:gen_result} demonstrates that GraphOpt learns a construction policy that effectively captures structural patterns in the observed network and constructs graphs with similar properties to the observed graph.
GraphOpt registers consistent and significantly superior performance across all datasets and against all baselines, which can perform well on some but not all metrics as they model specific statistics (except NetGAN).
BTER recovers clustering coefficient statistics well but struggles on others; DC-SBM recovers Max Degree better than others.
Further, \Cref{fig:coar_pol} demonstrates the ability of Graphopt to capture intrinsic properties of graph structure. Our stochastic policy ensures that generated graphs are not merely copies of the observed graph, which is further substantiated by link prediction experiments in Section 5.2. 
These performance characteristics are also visible for NetGAN, which performs well in general across all metrics and datasets.
However, our superior performance can be attributed to the following differences: (i) Our construction policy optimises a useful latent objective, whereas NetGAN's generator imitates the given graph by optimizing against an eventually uninformative discriminator;
(ii) Our use of GNN captures better structural properties to provide rich state information to the policy network, in contrast to LSTM based path processing in NetGAN;
(iii) GraphOpt allows construction of disconnected components, often found in most real-world graphs such as in these datasets.
Given these properties, we envision the use of GraphOpt as a graph constructor that ingests real-world graphs and generates synthetic versions to enrich graph repositories~ \citep{liu2020} with large-scale benchmark test sets. 

We provide more details on baselines/metrics used for experiments in \Cref{app:baselines} and \Cref{app:metrics} respectively. 

\vspace{-0.2cm}
\section{Conclusion}
In this work, we investigate a novel setting for learning over graphs that is motivated by the optimization perspective of graph formation in network science. 
Our novel optimization-based learning framework, GraphOpt, models graph formation as a sequential decision-making process, learns a forward model of graph construction, and discovers a latent objective that is  operationally equivalent to some underlying mechanism that could explain the formation of edges in observed graph. GraphOpt employs a novel combination of structured policy network, continuous latent action space and inverse reinforcement learning.
Empirically, GraphOpt discovers a latent objective and a robust stochastic policy that transfer across graphs with different characteristics,  exhibit competitive generalization for link prediction task and enable construction of graphs with similar properties as that of the observed graph.
We believe that our investigation of the optimization-based perspective on network formation stemming from the wider debate in network science literature and its implications for building sophisticated models to learn effectively from graph-structured observations, coupled with the versatility of our proposed approach, would benefit the graph learning community and open exciting avenues for future research.
\section*{Acknowledgements}
We thank the anonymous reviewers for critical feedback that helped us to improve the clarity and precision of our presentation. This work was supported by NSF IIS 1717916.  Part of the work done by Hongyuan Zha is supported by Shenzhen Institute of Artificial Intelligence and Robotics for Society, and Shenzhen Research Institute of Big Data.

\bibliography{citation}

\begin{thebibliography}{78}
\providecommand{\natexlab}[1]{#1}
\providecommand{\url}[1]{\texttt{#1}}
\expandafter\ifx\csname urlstyle\endcsname\relax
  \providecommand{\doi}[1]{doi: #1}\else
  \providecommand{\doi}{doi: \begingroup \urlstyle{rm}\Url}\fi

\bibitem[Abbeel \& Ng(2004)Abbeel and Ng]{alirl04}
Abbeel, P. and Ng, A.~Y.
\newblock Apprenticeship learning via inverse reinforcement learning.
\newblock In \emph{ICML}, 2004.

\bibitem[Adamic \& Glance(2005)Adamic and Glance]{AdaGla05}
Adamic, L.~A. and Glance, N.
\newblock The political blogosphere and the 2004 u.s. election: Divided they
  blog.
\newblock In \emph{Proceedings of the 3rd International Workshop on Link
  Discovery}, 2005.

\bibitem[Airoldi et~al.(2009)Airoldi, Blei, Fienberg, and Xing]{AirBleFieXin09}
Airoldi, E.~M., Blei, D.~M., Fienberg, S.~E., and Xing, E.~P.
\newblock Mixed membership stochastic blockmodels.
\newblock In \emph{Advances in Neural Information Processing Systems 21}. 2009.

\bibitem[Albert \& Barab\'asi(2002)Albert and Barab\'asi]{AlbBar02}
Albert, R. and Barab\'asi, A.-L.
\newblock Statistical mechanics of complex networks.
\newblock \emph{Rev. Mod. Phys.}, 2002.

\bibitem[Barab{\'a}si(2012)]{barabasi2012network}
Barab{\'a}si, A.-L.
\newblock Network science: Luck or reason.
\newblock \emph{Nature}, 2012.

\bibitem[Barab{\'a}si \& Albert(1999)Barab{\'a}si and
  Albert]{barabasi1999emergence}
Barab{\'a}si, A.-L. and Albert, R.
\newblock Emergence of scaling in random networks.
\newblock \emph{science}, 1999.

\bibitem[Barabási \& Pósfai(2016)Barabási and Pósfai]{barabasi2016network}
Barabási, A.-L. and Pósfai, M.
\newblock \emph{Network science}.
\newblock Cambridge University Press, 2016.

\bibitem[Bojchevski et~al.(2018)Bojchevski, Shchur, Z{\"u}gner, and
  G{\"u}nnemann]{BojShcZugCun18}
Bojchevski, A., Shchur, O., Z{\"u}gner, D., and G{\"u}nnemann, S.
\newblock Netgan: Generating graphs via random walks.
\newblock In \emph{Proceedings of the 35th International Conference on Machine
  Learning}, 2018.

\bibitem[Bojja et~al.(2018)Bojja, Alizadeh, and Viswanath]{graph2seq2018}
Bojja, S.~V., Alizadeh, M., and Viswanath, P.
\newblock Graph2seq: Scalable learning dynamics for graphs.
\newblock \emph{arXiv:1802.04948v3}, 2018.

\bibitem[Broido \& Clauset(2018)Broido and Clauset]{BroClau18}
Broido, A.~D. and Clauset, A.
\newblock Scale-free networks are rare.
\newblock In \emph{Nature Communications}, 2018.

\bibitem[Cao \& Kipf(2018)Cao and Kipf]{CaoKip18}
Cao, N.~D. and Kipf, T.
\newblock Molgan: An implicit generative model for small molecular graphs.
\newblock \emph{arXiv:1805.11973}, 2018.

\bibitem[Chakrabarti \& Faloutsos(2006)Chakrabarti and Faloutsos]{ChaFal06}
Chakrabarti, D. and Faloutsos, C.
\newblock Graph mining: Laws, generators, and algorithms.
\newblock \emph{ACM Comput. Surv.}, 2006.

\bibitem[Dai et~al.(2017)Dai, Khalil, Zhang, Dilkina, and
  Song]{DaiKhaYuyetal17}
Dai, H., Khalil, E.~B., Zhang, Y., Dilkina, B., and Song, L.
\newblock Learning combinatorial optimization algorithms over graphs.
\newblock In \emph{NIPS}, 2017.

\bibitem[Das et~al.(2018)Das, Dhuliawala, Zaheer, Vilnis, Durugkar,
  Krishnamurthy, Smola, and McCallum]{DasDhuZahVil18}
Das, R., Dhuliawala, S., Zaheer, M., Vilnis, L., Durugkar, I., Krishnamurthy,
  A., Smola, A., and McCallum, A.
\newblock Go for a walk and arrive at the answer: Reasoning over paths in
  knowledge bases using reinforcement learning.
\newblock 2018.

\bibitem[Dettmers et~al.(2018)Dettmers, Minervini, Stenetorp, and
  Riedel]{DetMinSteRie18}
Dettmers, T., Minervini, P., Stenetorp, P., and Riedel, S.
\newblock Convolutional 2d knowledge graph embeddings.
\newblock In \emph{AAAI}, 2018.

\bibitem[Dong et~al.(2017)Dong, Johnson, Xu, and Chawla]{DonJohXuCha17}
Dong, Y., Johnson, R.~A., Xu, J., and Chawla, N.~V.
\newblock Structural diversity and homophily: A study across more than one
  hundred big networks.
\newblock In \emph{Proceedings of the 23rd ACM SIGKDD International Conference
  on Knowledge Discovery and Data Mining}, 2017.

\bibitem[D'souza et~al.(2007)D'souza, Borgs, Chayes, Berger, and
  Kleinberg]{d2007emergence}
D'souza, R.~M., Borgs, C., Chayes, J.~T., Berger, N., and Kleinberg, R.~D.
\newblock Emergence of tempered preferential attachment from optimization.
\newblock \emph{Proceedings of the National Academy of Sciences}, 2007.

\bibitem[Erdos \& Renyi(1959)Erdos and Renyi]{ErdRen59}
Erdos, P. and Renyi, A.
\newblock On random graphs i.
\newblock \emph{Publicationes Mathematicae (Debrecen)}, 1959.

\bibitem[Fabrikant et~al.(2002)Fabrikant, Koutsoupias, and
  Papadimitriou]{fabrikant2002heuristically}
Fabrikant, A., Koutsoupias, E., and Papadimitriou, C.~H.
\newblock Heuristically optimized trade-offs: A new paradigm for power laws in
  the internet.
\newblock In \emph{International Colloquium on Automata, Languages, and
  Programming}, pp.\  110--122. Springer, 2002.

\bibitem[Finn et~al.(2016)Finn, Levine, and Abbeel]{finn2016guided}
Finn, C., Levine, S., and Abbeel, P.
\newblock Guided cost learning: Deep inverse optimal control via policy
  optimization.
\newblock In \emph{International Conference on Machine Learning}, pp.\  49--58,
  2016.

\bibitem[Fox et~al.(2016)Fox, Pakman, and Tishby]{Fox2016}
Fox, R., Pakman, A., and Tishby, N.
\newblock Taming the noise in reinforcement learning via soft updates.
\newblock In \emph{UAI}, 2016.

\bibitem[Geses et~al.(2019)Geses, Biswas, Alam, and Sack]{Gen2019}
Geses, G.~A., Biswas, R., Alam, M., and Sack, H.
\newblock A survey on knowledge graph embeddings with literals: Which model
  links better literal-ly?
\newblock \emph{arXiv:1910.12507}, 2019.

\bibitem[Grover \& Leskovec(2016)Grover and Leskovec]{Grover2016}
Grover, A. and Leskovec, J.
\newblock Node2vec: Scalable feature learning for networks.
\newblock In \emph{KDD}, 2016.

\bibitem[Haarnoja et~al.(2017)Haarnoja, Tang, Abbeel, and Levine]{haarnoja17}
Haarnoja, T., Tang, H., Abbeel, P., and Levine, S.
\newblock Reinforcement learning with deep energy-based policies.
\newblock In \emph{ICML}, 2017.

\bibitem[Haarnoja et~al.(2018)Haarnoja, Zhou, Hartikainen, Tucker, Ha, Tan,
  Kumar, Zhu, Gupta, Abbeel, and Levine]{haarnoja2018soft}
Haarnoja, T., Zhou, A., Hartikainen, K., Tucker, G., Ha, S., Tan, J., Kumar,
  V., Zhu, H., Gupta, A., Abbeel, P., and Levine, S.
\newblock Soft actor-critic algorithms and applications.
\newblock \emph{arxiv:1812.05905}, 2018.

\bibitem[Hamilton et~al.(2017)Hamilton, Ying, and Leskovec]{HamYinLes17}
Hamilton, W.~L., Ying, R., and Leskovec, J.
\newblock Representation learning on graphs: Methods and applications.
\newblock \emph{arXiv:1709.05584}, 2017.

\bibitem[Ho \& Ermon(2016)Ho and Ermon]{ho2016generative}
Ho, J. and Ermon, S.
\newblock Generative adversarial imitation learning.
\newblock In \emph{Advances in Neural Information Processing Systems}, pp.\
  4565--4573, 2016.

\bibitem[Karrer \& Newman(2010)Karrer and Newman]{KarNew10}
Karrer, B. and Newman, M.
\newblock Stochastic blockmodels and community structure in networks.
\newblock \emph{Physical Review E}, 2010.

\bibitem[Kipf \& Welling(2016)Kipf and Welling]{kipf2016variational}
Kipf, T.~N. and Welling, M.
\newblock Variational graph auto-encoders.
\newblock \emph{arXiv preprint arXiv:1611.07308}, 2016.

\bibitem[Kok \& Domingos(2007)Kok and Domingos]{KokDom07}
Kok, S. and Domingos, P.
\newblock Statistical predicate invention.
\newblock In \emph{ICML}, 2007.

\bibitem[Konda \& Tsitsiklis(2000)Konda and Tsitsiklis]{konda2000actor}
Konda, V.~R. and Tsitsiklis, J.~N.
\newblock Actor-critic algorithms.
\newblock In \emph{Advances in neural information processing systems}, pp.\
  1008--1014, 2000.

\bibitem[Kumar et~al.(2000)Kumar, Raghavan, Rajagopalan, Sivakumar, Tompkins,
  and Upfal]{kumar2000web}
Kumar, R., Raghavan, P., Rajagopalan, S., Sivakumar, D., Tompkins, A., and
  Upfal, E.
\newblock The web as a graph.
\newblock In \emph{Proceedings of the nineteenth ACM SIGMOD-SIGACT-SIGART
  symposium on Principles of database systems}, pp.\  1--10, 2000.

\bibitem[Leskovec et~al.(2007)Leskovec, Kleinberg, and Faloutsos]{LesKleFal07}
Leskovec, J., Kleinberg, J., and Faloutsos, C.
\newblock Graph evolution: Densification and shrinking diameters.
\newblock \emph{ACM Trans. Knowl. Discov. Data}, 2007.

\bibitem[Leskovec et~al.(2010)Leskovec, Chakrabarti, Kleinberg, Faloutsos, and
  Ghahramani]{LesChaKleFal10}
Leskovec, J., Chakrabarti, D., Kleinberg, J., Faloutsos, C., and Ghahramani, Z.
\newblock Kronecker graphs: An approach to modeling networks.
\newblock \emph{J. Mach. Learn. Res.}, 2010.

\bibitem[Levine(2018)]{Levine2018}
Levine, S.
\newblock Reinforcement learning and control as probabilistic inference:
  Tutorial and review.
\newblock \emph{arxiv:1805.00909}, 2018.

\bibitem[Li et~al.(2018{\natexlab{a}})Li, Vinyals, Dyer, Pascanu, and
  Battaglia]{li2018learning}
Li, Y., Vinyals, O., Dyer, C., Pascanu, R., and Battaglia, P.
\newblock Learning deep generative models of graphs.
\newblock \emph{arXiv preprint arXiv:1803.03324}, 2018{\natexlab{a}}.

\bibitem[Li et~al.(2018{\natexlab{b}})Li, Chen, and Koltun]{LiCheKol18}
Li, Z., Chen, Q., and Koltun, V.
\newblock Combinatorial optimization with graph convolutional networks and
  guided tree search.
\newblock In \emph{NeurIPS}, 2018{\natexlab{b}}.

\bibitem[Lillicrap et~al.(2015)Lillicrap, Hunt, Pritzel, Heess, Erez, Tassa,
  Silver, and Wierstra]{lillicrap2015continuous}
Lillicrap, T.~P., Hunt, J.~J., Pritzel, A., Heess, N., Erez, T., Tassa, Y.,
  Silver, D., and Wierstra, D.
\newblock Continuous control with deep reinforcement learning.
\newblock \emph{arXiv preprint arXiv:1509.02971}, 2015.

\bibitem[Lin et~al.(2018)Lin, Socher, and Xiong]{Xi2018}
Lin, X.~V., Socher, R., and Xiong, C.
\newblock Multi-hop knowledge graph reasoning with reward shaping.
\newblock \emph{arxiv:1808.10568}, 2018.

\bibitem[Liu et~al.(2020)Liu, Hu, Zitnik, and Leskovec]{liu2020}
Liu, B., Hu, W., Zitnik, M., and Leskovec, J.
\newblock Open graph benchmark.
\newblock \emph{To appear}, 2020.

\bibitem[Liu et~al.(2016)Liu, Cheung, Li, and Liao]{LiuCheLiLia16}
Liu, L., Cheung, W.~K., Li, X., and Liao, L.
\newblock Aligning users across social networks using network embedding.
\newblock In \emph{Proceedings of the Twenty-Fifth International Joint
  Conference on Artificial Intelligence}, 2016.

\bibitem[McCallum et~al.(2000)McCallum, Nigam, Rennie, and
  Seymore]{McNigRenSy00}
McCallum, A.~K., Nigam, K., Rennie, J., and Seymore, K.
\newblock Automating the construction of internet portals with machine
  learning.
\newblock \emph{Information Retrieval}, 2000.

\bibitem[Mnih et~al.(2015)Mnih, Kavukcuoglu, Silver, Rusu, Veness, Bellemare,
  Graves, Riedmiller, Fidjeland, Ostrovski, et~al.]{mnih2015human}
Mnih, V., Kavukcuoglu, K., Silver, D., Rusu, A.~A., Veness, J., Bellemare,
  M.~G., Graves, A., Riedmiller, M., Fidjeland, A.~K., Ostrovski, G., et~al.
\newblock Human-level control through deep reinforcement learning.
\newblock \emph{Nature}, 518\penalty0 (7540):\penalty0 529, 2015.

\bibitem[Newman(2010)]{newman2010networks}
Newman, M.
\newblock \emph{Networks: An Introduction}.
\newblock OUP Oxford, 2010.

\bibitem[Ng \& Russell(2000)Ng and Russell]{ng2000}
Ng, A.~Y. and Russell, S.
\newblock Algorithms for inverse reinforcement learning.
\newblock In \emph{in Proc. 17th International Conf. on Machine Learning}, pp.\
   663--670. Morgan Kaufmann, 2000.

\bibitem[Papadopoulos et~al.(2012)Papadopoulos, Kitsak, Serrano, Bogun{\'a},
  and Krioukov]{papadopoulos2012popularity}
Papadopoulos, F., Kitsak, M., Serrano, M.~{\'A}., Bogun{\'a}, M., and Krioukov,
  D.
\newblock Popularity versus similarity in growing networks.
\newblock \emph{Nature}, 2012.

\bibitem[Paula et~al.(2018)Paula, Richards-Shubik, and Tamer]{PauRicTam18}
Paula, A.~D., Richards-Shubik, S., and Tamer, E.
\newblock Identifying preferences in networks with bounded degree.
\newblock \emph{Econometrica}, 2018.

\bibitem[Rawlik et~al.(2013)Rawlik, Toussaint, and Vijayakumar]{Rawlik2013}
Rawlik, K., Toussaint, M., and Vijayakumar, S.
\newblock On stochastic optimal control and reinforcement learning by
  approximate inference (extended abstract).
\newblock In \emph{IJCAI}, 2013.

\bibitem[Robins et~al.(2007)Robins, Pattison, Kalish, and
  Lusher]{RobPatKalLus07}
Robins, G., Pattison, P., Kalish, Y., and Lusher, D.
\newblock An introduction to exponential random graph (p*) models for social
  networks.
\newblock \emph{Social Networks}, 2007.

\bibitem[{Scarselli} et~al.(2009){Scarselli}, {Gori}, {Tsoi}, {Hagenbuchner},
  and {Monfardini}]{Sca09}
{Scarselli}, F., {Gori}, M., {Tsoi}, A.~C., {Hagenbuchner}, M., and
  {Monfardini}, G.
\newblock The graph neural network model.
\newblock \emph{IEEE Transactions on Neural Networks}, 2009.

\bibitem[Schulman et~al.(2017)Schulman, Wolski, Dhariwal, Radford, and
  Klimov]{SchWolDhaRadKli17}
Schulman, J., Wolski, F., Dhariwal, P., Radford, A., and Klimov, O.
\newblock Proximal policy optimization algorithms.
\newblock \emph{arXiv:1707.06347}, 2017.

\bibitem[Sen et~al.(2008)Sen, Namata, Bilgic, Getoor, Galligher, and
  Eliassi-Rad]{SenNamBilGetGalEli2}
Sen, P., Namata, G., Bilgic, M., Getoor, L., Galligher, B., and Eliassi-Rad.
\newblock Collective classification in network data.
\newblock \emph{Ai Magazine}, 2008.

\bibitem[Seshadhri et~al.(2012)Seshadhri, Kolda, and Pinar]{SesKolPin16}
Seshadhri, C., Kolda, T.~G., and Pinar, A.
\newblock Community structure and scale-free collections of erdos-renyi graphs.
\newblock \emph{Physical Review E}, 2012.

\bibitem[Shen et~al.(2018)Shen, Chen, Huang, Guo, and Gao]{shenmwalk18}
Shen, Y., Chen, J., Huang, P.-S., Guo, Y., and Gao, J.
\newblock M-walk: Learning to walk over graphs using monte carlo tree search.
\newblock In \emph{NeurIPS}, 2018.

\bibitem[Silver et~al.(2016)Silver, Huang, Maddison, Guez, Sifre, Van
  Den~Driessche, Schrittwieser, Antonoglou, Panneershelvam, Lanctot,
  et~al.]{silver2016mastering}
Silver, D., Huang, A., Maddison, C.~J., Guez, A., Sifre, L., Van Den~Driessche,
  G., Schrittwieser, J., Antonoglou, I., Panneershelvam, V., Lanctot, M.,
  et~al.
\newblock Mastering the game of go with deep neural networks and tree search.
\newblock \emph{nature}, 529\penalty0 (7587):\penalty0 484, 2016.

\bibitem[Silver et~al.(2017)Silver, Schrittwieser, Simonyan, Antonoglou, Huang,
  Guez, Hubert, Baker, Lai, Bolton, Chen, Lillicrap, Hui, Sifre, van~den
  Driessche, Graepel, and Hassabis]{silver2017mastering}
Silver, D., Schrittwieser, J., Simonyan, K., Antonoglou, I., Huang, A., Guez,
  A., Hubert, T., Baker, L., Lai, M., Bolton, A., Chen, Y., Lillicrap, T., Hui,
  F., Sifre, L., van~den Driessche, G., Graepel, T., and Hassabis, D.
\newblock Mastering the game of go without human knowledge.
\newblock \emph{Nature}, 2017.

\bibitem[Simonovsky \& Komodakis(2018)Simonovsky and Komodakis]{SimKom18}
Simonovsky, M. and Komodakis, N.
\newblock Graphvae: Towards generation of small graphs using variational
  autoencoders.
\newblock \emph{arxiv:1802.03480}, 2018.

\bibitem[Singh \& Lio(2019)Singh and Lio]{SinLio19}
Singh, V. and Lio, P.
\newblock Towards probabilistic generative models harnessing graph neural
  networks for disease-gene prediction.
\newblock \emph{arxiv:1907.05628}, 2019.

\bibitem[Sutton \& Barto(2018)Sutton and Barto]{sutton2018reinforcement}
Sutton, R.~S. and Barto, A.~G.
\newblock \emph{Reinforcement learning: An introduction}.
\newblock MIT press, 2018.

\bibitem[Sutton et~al.(2000)Sutton, McAllester, Singh, and
  Mansour]{sutton2000policy}
Sutton, R.~S., McAllester, D.~A., Singh, S.~P., and Mansour, Y.
\newblock Policy gradient methods for reinforcement learning with function
  approximation.
\newblock In \emph{Advances in neural information processing systems}, pp.\
  1057--1063, 2000.

\bibitem[Torabi et~al.(2018)Torabi, Warnell, and Stone]{torabi2018behavioral}
Torabi, F., Warnell, G., and Stone, P.
\newblock Behavioral cloning from observation.
\newblock In \emph{Proceedings of the 27th International Joint Conference on
  Artificial Intelligence}, pp.\  4950--4957, 2018.

\bibitem[Toussaint(2009)]{Toussaint2009}
Toussaint, M.
\newblock Robot trajectory optimization using approximate inference.
\newblock In \emph{ICML}, 2009.

\bibitem[Toutanova et~al.(2015)Toutanova, Chen, Pantel, Poon, Choudhury, and
  Gamon]{TotChePan15}
Toutanova, K., Chen, D., Pantel, P., Poon, H., Choudhury, P., and Gamon, M.
\newblock Representing text for joint embedding of text and knowledge bases.
\newblock In \emph{EMNLP}, 2015.

\bibitem[Trouillon et~al.(2016)Trouillon, Welbl, Riedel, Gaussier, and
  Bouchard]{Trouillon2016}
Trouillon, T., Welbl, J., Riedel, S., Gaussier, E., and Bouchard, G.
\newblock Complex embeddings for simple link prediction.
\newblock In \emph{ICML}, 2016.

\bibitem[V{\'a}zquez et~al.(2003)V{\'a}zquez, Flammini, Maritan, and
  Vespignani]{vazquez2003modeling}
V{\'a}zquez, A., Flammini, A., Maritan, A., and Vespignani, A.
\newblock Modeling of protein interaction networks.
\newblock \emph{Complexus}, 1\penalty0 (1):\penalty0 38--44, 2003.

\bibitem[Wang et~al.(2017)Wang, Wang, Wang, Zhao, Zhang, Zhang, Xie, and
  Guo]{wang2017graphgan}
Wang, H., Wang, J., Wang, J., Zhao, M., Zhang, W., Zhang, F., Xie, X., and Guo,
  M.
\newblock Graphgan: Graph representation learning with generative adversarial
  nets.
\newblock \emph{1711.08267}, 2017.

\bibitem[Wang et~al.(2018)Wang, Liao, Ba, and Fidler]{Wang18}
Wang, T., Liao, R., Ba, J., and Fidler, S.
\newblock Nervenet: Learning structured policy via graph neural networks.
\newblock In \emph{ICLR}, 2018.

\bibitem[Williams(1992)]{Will92}
Williams, R.~J.
\newblock Simple statistical gradient-following algorithms for connectionist
  reinforcement learning.
\newblock \emph{Machine Learning}, 1992.

\bibitem[Xiao et~al.(2016)Xiao, Huang, and Zhu]{XiaHuaZhu16}
Xiao, H., Huang, M., and Zhu, X.
\newblock Transg : A generative model for knowledge graph embedding.
\newblock In \emph{Proceedings of the 54th Annual Meeting of the Association
  for Computational Linguistic}, 2016.

\bibitem[Xiong et~al.(2017)Xiong, Hoang, and Wang]{xiong2017deeppath}
Xiong, W., Hoang, T., and Wang, W.~Y.
\newblock Deeppath: A reinforcement learning method for knowledge graph
  reasoning.
\newblock \emph{arXiv preprint arXiv:1707.06690}, 2017.

\bibitem[Yang et~al.(2017)Yang, Yang, and Cohen]{NeuralLP17}
Yang, F., Yang, Z., and Cohen, W.~W.
\newblock Differentiable learning of logical rules for knowledge base
  completion.
\newblock \emph{arxiv:1702.08367}, 2017.

\bibitem[You et~al.(2018{\natexlab{a}})You, Liu, Ying, Pande, and
  Leskovec]{YouLiuYinPanLes18}
You, J., Liu, B., Ying, R., Pande, V., and Leskovec, J.
\newblock Graph convolutional policy network for goal directed molecule
  generation.
\newblock In \emph{NIPS}, 2018{\natexlab{a}}.

\bibitem[You et~al.(2018{\natexlab{b}})You, Ying, Ren, Hamilton, and
  Leskovec]{you2018graphrnn}
You, J., Ying, R., Ren, X., Hamilton, W.~L., and Leskovec, J.
\newblock Graphrnn: Generative realistic graphs with deep auto-regressive
  mdoels.
\newblock \emph{arXiv preprint arXiv:1802.08773}, 2018{\natexlab{b}}.

\bibitem[Zhang \& Chen(2018)Zhang and Chen]{seal2018}
Zhang, M. and Chen, Y.
\newblock Link prediction based on graph neural networks.
\newblock \emph{arxiv:1802.09691}, 2018.

\bibitem[Zhang et~al.(2017)Zhang, Yao, and Sun]{ZhaYaoSun18}
Zhang, S., Yao, L., and Sun, A.
\newblock Deep learning based recommender system: {A} survey and new
  perspectives.
\newblock \emph{arXiv:1707.07435}, 2017.

\bibitem[Zhang et~al.(2018)Zhang, Cui, and Zhu]{ZhaCuiZhu18}
Zhang, Z., Cui, P., and Zhu, W.
\newblock Deep learning on graphs: A survey.
\newblock \emph{arXiv:1812.04202}, 2018.

\bibitem[Ziebart(2010)]{ziebart2010modelingB}
Ziebart, B.~D.
\newblock \emph{Modeling Purposeful Adaptive Behavior with the Principle of
  Maximum Causal Entropy}.
\newblock PhD thesis, Machine Learning Department, Carnegie Mellon University,
  2010.

\bibitem[Ziebart et~al.(2008)Ziebart, Maas, Bagnell, and Dey]{ZieMaaBagdey08}
Ziebart, B.~D., Maas, A., Bagnell, J., and Dey, A.~K.
\newblock Maximum entropy inverse reinforcement learning.
\newblock In \emph{AAAI}, 2008.

\end{thebibliography}
\bibliographystyle{icml2020}

\appendix
\onecolumn

\section{Gradient Updates for GraphOpt Algorithm}

In this section,  we provide details on the gradient update algorithm:

\label{app:gradient-updates}

\begin{algorithm}[ht!]
   \caption{Parameter Update}
   \label{alg:TrainPolicy}
   \begin{algorithmic}[1]
     \Procedure{train\_policy}{$B, \psi, \phi, \theta, \bar{\psi}, \omega$}
     \For{each gradient step}
      	\State Sample mini-batch of $\mathcal{M}$ transitions
      	\State $\lbrace (s_t,a_t,r_t(s_t,a_t),s_{t+1}) \rbrace$ from $B$
    	\State Compute $\hat{\nabla}_\psi J_V(\psi), \hat{\nabla}_\theta J_Q(\theta), \hat{\nabla}_\phi J_\pi(\phi)$~\text{using}
    	\State Eq.~\ref{eq:q-v},\ref{eq:grad-q}, \ref{eq:grad-pi} and compute $\nabla_{\omega}J$ for repr. network
        \State Update the parameters based on following:
        \State $\psi \leftarrow \psi - \lambda_V \hat{\nabla}_\psi J_V(\psi)$
        \State $\theta_i \leftarrow \theta_i - \lambda_Q \hat{\nabla}_\theta J_Q(\theta_i)~~i \in \{1,2\}$
        \State $\phi \leftarrow \phi - \lambda_\pi \hat{\nabla}_\phi J_\pi(\phi)$
        \State $\bar{\psi} \leftarrow l\bar{\psi} + (1 - l)\bar{\psi}$
        \State $\omega \leftarrow \omega - \lambda_{emb} \hat{\nabla}_\omega J_{emb}(\omega)$
    \EndFor
     \EndProcedure
  \end{algorithmic}
\end{algorithm}

For completeness, we present the gradients of each objective below, however, they are adopted from~\cite{haarnoja2018soft} and we encourage interested readers to refer to the original manuscript for more details:
\begin{enumerate}
    \item {\bf Soft Q-function} is trained to minimize the Bellman residual:
    
    \begin{equation}
        \label{eq:bellman-residual}
        J_Q(\theta) = \mathbb{E}_{(s_t,a_t) \sim B} \left[\frac{1}{2}\left( Q_\theta(s_t,a_t) - \hat{Q}(s_t,a_t)\right)^2\right]
    \end{equation}
    where,
    \begin{equation}
        \label{eq:q-v}
        \hat{Q}(s_t,a_t) = r(s_t, a_t) + \gamma \mathbb{E}{s_{t+1}\sim p}[V_{\bar{\psi}}(s_{t+1})]
    \end{equation}
    where $V_{\bar{\psi}}$ is a target value network and $\bar{\psi}$ is an exponentially moving average of the value network weights for stabilizing training.
    The gradients for Eq.~\ref{eq:bellman-residual} is given by:
    \begin{equation}
    \label{eq:grad-q}
        \begin{split}
            \hat{\nabla}_\theta J_Q(\theta) &= \nabla_\theta Q_\theta(s_t, a_t)(Q_\theta(s_t, a_t) \\&- r(s_t, a_t) - \gamma V_{\bar{\psi}}(s_{t+1}))
        \end{split}
    \end{equation}
    
    \item {\bf Policy Network} is trained  using the following objective function:
    \begin{equation}
    \label{eq:policy-objective}
        \begin{split}
            J_\pi(\phi) &= \mathbb{E}_{s_t \sim B, \epsilon_t \sim \mathcal{N}}[\log\pi_\phi(f_\phi(\epsilon_t;s_t)|s_t) \\&-
            Q_\theta(s_t, f_\phi(\epsilon_t;s_t))]
        \end{split}
    \end{equation}
    where $f_\phi$ is the transformation applied to the policy network to induce reparameterization that helps in building a low variance estimator.
    The gradient for Eq.~\ref{eq:policy-objective} is then given by:
    \begin{equation}
    \label{eq:grad-pi}
        \begin{split}
            \hat{\nabla}_\phi J_\pi(\phi) &= \nabla_\phi \log \pi_\phi (a_t|s_t) \\&+ \nabla_\phi f_\phi(\epsilon_t;s_t)\left(\nabla_{a_t}\log\pi_\phi(a_t|s_t) \right. \\&- \left.  \nabla_{a_t}Q(s_t, a_t)
            \right)
        \end{split}
    \end{equation}
\end{enumerate}

\newpage

\section{More Related Work}

\label{app:relatedworks}

{\bf Reinforcement Learning for Graphs.} Recent advancements in deep learning techniques over graph structured data~\cite{ZhaCuiZhu18,HamYinLes17} 
and progress in deep RL~\cite{lillicrap2015continuous, SchWolDhaRadKli17} has stimulated increased interest in casting the general task of learning over graphs into a sequential decision process, whereby actions correspond to the discrete set of nodes to be sequentially connected to a partially constructed graph.
This procedure optimizes task-specific objectives in the form of the reward function. 
Specifically, deep reinforcement learning has been used in three major learning paradigms: \textit{Learning to Generate} \citep{YouLiuYinPanLes18} proposes a generative model for molecular structure using a graph convolutional policy network to optimize a domain specific reward that captures various properties of molecules. \textit{Learning to Walk} \citep{xiong2017deeppath, DasDhuZahVil18, shenmwalk18} uses RL for tasks of link prediction and QA over Knowledge Graphs. 
The key goal in these works is to find an optimal path from a query entity to a target answer entity. \textit{Learning to Optimize} \citep{DaiKhaYuyetal17,LiCheKol18,graph2seq2018}: builds a combination of graph neural networks and RL to learn optimization algorithms for NP-hard 
problems (e.g. MaxCut).

{\bf Maximum Entropy Reinforcement Learning.} Deep Reinforcement Learning~~\cite{mnih2015human,silver2017mastering,silver2016mastering}, specifically Actor-critic algorithms~\cite{konda2000actor} has recently achieved great success in a variety of tasks that require search over combinatorial space and have inspired many new architectures that can be broadly categorized into: on-policy algorithms~\cite{sutton2000policy, SchWolDhaRadKli17} that build on standard on-policy policy gradients and off-policy algorithms~\cite{lillicrap2015continuous} that use off-policy samples from replay buffer. 
Both categories exhibit trade-off between sample complexity and stability with on-policy algorithms being more stable while off-policy counterpart being more sample efficient.
We build our framework on recently proposed Soft Actor-Critic (SAC)~\cite{haarnoja2018soft} algorithm that has been shown to be both sample efficient and stable. SAC falls in the category of Maximum entropy based algorithms~\cite{ZieMaaBagdey08, Toussaint2009, Rawlik2013, Fox2016, Levine2018} that are based on maximum entropy learning~\cite{ziebart2010modelingB} and have been shown to be robust in the face of model and estimation errors while improving exploration~\cite{haarnoja17}.
 \textit{\bf Inverse Reinforcement Learning (IRL)} --- Methods in IRL \citep{ng2000} and imitation learning ~\citep{alirl04} seek to recover a reward function given measurements of near-optimal expert trajectories. The maximum entropy IRL framework \citep{ZieMaaBagdey08} leads to a sample-based method that learns a neural network approximation of the reward, without requiring knowledge of the MDP transition function \citep{finn2016guided}.

{\bf Deep Generative Models of Graph Generation.} Recently, there have been significant research efforts in building deep generative models of graph generation as they allow to effectively capture complex structural properties observed in a graph and use that information to output realistic graphs. Most of these works can be broadly categorized into two classes: (i) Methods that learn from collection of graphs (e.g. DeepGMG~\cite{li2018learning}, GraphRNN~\cite{you2018graphrnn}, GCPN~\cite{YouLiuYinPanLes18}) and (ii) Methods that learn from a single graph (e.g VGAE~\cite{kipf2016variational}, GraphGan~\cite{wang2017graphgan}, MolGAN~\cite{CaoKip18}, NetGan~\cite{BojShcZugCun18}). As discussed in the main paper, our current approach falls into the second category.
DeepGMG builds probabilistic model where the partially generated graph is encoded by the graph neural network (GNN) and the representation is used to make decision of constructing next node or edge. However, it suffers from scalability issues. GraphRNN solves this problem by proposing an auto-regressive model of graph generation, wherein the focus is on generating sequence of adjacency vectors that be mapped to graph structure. It employs hierarchical recurrent architecture to encode the historical path information. GCPN combines GCN with RL and learns a deep generative model using an objective specific to domain of chemistry. In the second category, 
methods like GraphGan and GVAE are implicit models but their main focus is to learn graph representations and hence perform weakly on generation tasks and have limited scalability. NetGan is a recently proposed implict graph generator model exhibiting generalization properties. However, unlike GraphOpt, NetGAN optimizes a GAN-based objective which converge to an uninformative discriminator, thereby not useful for transfer, which in contrast is a key objective of our approach.

\newpage
\section{Additional Details on Experiments}
\label{app:implementation}

\subsection{Datasets}
\label{app:datasets}

Table~\ref{tab:datastat_gen} provide statistics and reference to the dataset used for Graph Construction experiments. Table~\ref{tab:pred_graph1} provides dataset statistics for non-relational datasets and Table~\ref{tab:pred_graph2} provides dataset statistics for relational datasets, both used for link prediction.

\begin{table}[h]
  \centering
  \caption{Dataset Statistics for Construction Experiments}
  \resizebox{0.6\textwidth}{!}{
    \begin{tabular}{c|ccccc}
    \toprule
    \textbf{Graph} & \textbf{Nodes} & \textbf{Edges} & \textbf{Density} & \textbf{Avg. Degree} & \textbf{Source}\\
    \toprule
    Barabasi-Albert (BA) & 100   & 384 & 0.0384 & 7.68  & Synthetic\\
    Erdos-Renyi  & 500  & 6152 & 0.02461 & 24.608  & Synthetic\\
    Political Blogs & 1224 &  16718 & 0.0127 & 27.316 & \cite{AdaGla05}\\
    CORA-ML & 2995  & 8158  & 0.001 & 3.898 & \cite{McNigRenSy00}\\
    PubMed & 19717 & 44327 & 0.00011 & 4.496  & \cite{SenNamBilGetGalEli2}\\
    CiteSeer & 3327  & 4676  & 0.00042 & 2.811 & \cite{SenNamBilGetGalEli2}\\
    \bottomrule
    \end{tabular}}%
  \label{tab:datastat_gen}%
\end{table}%

\begin{table}[h]
  \centering
  \caption{Social, Metabolic and Citation Graphs for Link Prediction}
  \resizebox{0.6\textwidth}{!}{
    \begin{tabular}{c|ccccc}
    \toprule
    \textbf{Graph} & \textbf{Nodes} & \textbf{Edges} & \textbf{Density} & \textbf{Average Deg.} & \textbf{Source}\\
    \toprule
    Cora-ML & 2995  & 8158  & 0.001 & 3.898  & \cite{McNigRenSy00}\\
    Political Blogs & 1224  & 16718 & 0.0112 & 27.316 & \cite{AdaGla05}\\
    E. Coli & 1805  & 14660 & 0.0045 & 12.55 & \cite{seal2018}\\
    \bottomrule
    \end{tabular}}%
  \label{tab:pred_graph1}
  \hfill
\end{table}

\begin{table}[h]
  \centering
  \caption{Knowledge Graphs for Link Prediction}
  \resizebox{0.6\textwidth}{!}{
    \begin{tabular}{c|cccccc}
    \toprule
    \textbf{Graph} & \textbf{Nodes} & \textbf{Egdes} & \textbf{Relations} & \textbf{Density} & \textbf{Average Deg.} & \textbf{Source}\\
    \toprule
    Kinship & 104   & 10686 & 25    & 0.9879 & 82.15 & \cite{KokDom07}\\
    FB15k-237 & 14541 & 310116 & 237   & 0.0013 & 19.74 & \cite{TotChePan15}\\
    WN18RR & 40943   & 93003  & 11     & 0.00005 & 2.19 & \cite{DetMinSteRie18}\\
    \bottomrule
    \end{tabular}}%
  \label{tab:pred_graph2}%
\end{table}%

\subsection{Baselines}
\label{app:baselines}

As we learn from single graph as an observation, we compare with the state-of-art baselines that can operate in this setting for a fair comparison. Below we provide references and some details on each baseline that we compare with:

\begin{itemize}
    \item For graph construction experiments, we compare against two traditional generators: Degree Corrected Stochastic Block Model (DC-SBM)~\cite{KarNew10} that extends classical stochastic block models to account for heterogeneous degree of the vertices and block two level Erdos-Reniy (BTER) random graph model~\cite{SesKolPin16} that generates graph with dense subgraphs, each being an ER graph in itself and has a degree preserving property. Next, we use Variational Graph Autoencoder (VGAE)~\cite{kipf2016variational} that uses an autoencoding mechanism to learn node embeddings. Being an autorecnoder, it has the capacity to generate graphs and hence serves as a representative node embedding approach for graph generation. Finally, NetGAN~\cite{BojShcZugCun18} is the state-of-art implicit model of graph generation that learns to mimic network patterns observed in a graph by building a probabilistic models of random walks over the graph. NetGAN employs a GAN based objective where the discriminator attempts to distinguish the generated random walks from the observations.

    \item For link prediction experiments in non-relational domains, we compare again with NetGan and GVAE. NetGan is an implicit model of graph generation that generalizes to the task of link prediction which is also the property of our model and hence provides a strong baseline for our work. Further, GVAE being a node embedding approach, is naturally suitable for link prediction. Next, we compare Node2Vec~\cite{Grover2016}, a classical node embedding approach based on random walks. Finally, we compare with a strong state-of-art link prediction baseline, SEAL~\cite{seal2018}, that uses a graph neural network based architecture to perform link prediction. It is important to note that SEAL employs a task specific architecture and task-specific objective and hence provide a strong baseline to compare with.
    
    \item For relational link prediction on knowledge graphs, we cover a spectrum of approaches through their representative baselines. ConvE~\cite{DetMinSteRie18} is the state-of-art embedding based relational learning approach that has been shown to outperform most of earlier baselines in relational learning over knowledge literature~\cite{Gen2019}.  We consider two Rl based method, Minerva~\cite{DasDhuZahVil18} and Reward Shaping~\cite{Xi2018}. Minerva employs an LSTM encoder to encode the path information into the state representation and then solve a Markov Decision Process using policy gradients more specifically by REINFORCE algorithm~\cite{Will92}. Reward Shaping advances the methods in MINERVA to build a strong state-of-art baseline. However, it is again important to note both ConvE and Reward Shaping are dedicated baselines with task specific architecture and objective analogous to SEAL above (Reward shaping specifically shapes the reward using embedding model which essentially makes it very close to link prediction objective). On the other hand, MINVERVA can be considered somewhat closer in spirit to our work as it uses more general form of reward (+1/-1) which is also sparse while still using task-specific architecture. Finally, we also compare with NeuralLP~\cite{NeuralLP17} that introduces a differential rule learning system using symbolic operaters to learn logic rules to perform reasoning.
    
\end{itemize}

\subsection{Evaluation Protocol for Link Prediction using Learned Embeddings in \Cref{tab:linkpred_policy_nonr} and \Cref{tab:linkpred_policy_r}}
\label{app:evaluation-protocol}
 Our state encoder learns to represent the nodes into low dimensional representation. This experiment evaluates how well the encoder network was learned by testing the ability of embeddings learned by the encoder itself to perform link prediction. Once the model is trained, we pass the training graph through the encoder to get the final trained embedding of nodes. For non-relational case, we use dot product based embedding similarity criterion and label edges as before. For relational case, we use score function similar to ComplEx~\cite{Trouillon2016} and rank the entities as before. 

\subsection{GraphOpt Implementation}
\label{app:impl}

We closely follow SAC's architecture in using two Q-functions for efficient learning and parameterizing them with standard multi-layered perceptron. Similarly, the reward function $\Rcal$ is parameterized by standard multi-layer perceptron. As described in Section 3, we use GNN based policy network. We use the rlkit library\footnote{https://github.com/vitchyr/rlkit} available online and adapt it for our framework. In general, SAC does not have any aggressive hyper-parameters that needs to be tuned -- Table~\ref{tab:hyperparam} provides complete list of hyper-parameters with their reasonable default values that were used for maximum number of experiments and here we describe the ones that were mostly tuned using validation performance.

For evaluation, SAC provides two choices to use either deterministic or stochastic and we use stochastic for our evaluation as stochastic action is an important factor in our work. We tune the number of epochs ranging from 10000-50000 epochs depending on the graph environment. For inverse learning, we vary reward iterations from 30-300. Most of other hyper-parameters remain fairly constant across the environments. For state encoder, we tune prop-round between 2-4 and found 2 to be best choice. term\_threshold ranges from 100-500 based on dataset and node emebeding size is tuned in the range $\lbrace 32, 64, 128, 256, 512 \rbrace$. In the general parameters, net\_size represents the size of the hidden units of MLP that were used for policy, Q, representation and reward networks. We tune this parameter again in the range $\lbrace 32, 64, 128, 256, 512 \rbrace$ based on the environment.

All the experiments were conducted on Intel Xeon CPU V4 \@ 2.20 GHZ with 64 GB memory and Nvidia GeForce 1080 GPU.

\begin{table*}[t]
  \centering
  \caption{Hyper Parameter Configuration Table}
    \begin{tabular}{lr|lr}
    \toprule
    \multicolumn{4}{l}{\textbf{GraphOpt Algorithm}} \\
    \midrule
    \textbf{HyperParameters} & \textbf{Default Values} & \textbf{HyperParameters} & \textbf{Default Values} \\
    num\_epochs & 10000 & reward\_lr & 0.01 \\
    num\_steps\_per\_epoch & 500   & soft\_target\_tau & 0.001 \\
    num\_steps\_per\_eval & 1000  & policy\_lr & 1.00E-04 \\
    num\_steps\_before\_training\_online & 25    & qf\_lr & 1.00E-03 \\
    replay\_buffer\_size & 100000 & optimizer & Adam \\
    batch\_size & 128   & train\_policy\_with\_reprarameterization & TRUE \\
    max\_path\_length & 1000  & eval\_deterministic & FALSE \\
    discount & 0.99  & use\_automatic\_entropy\_tuning & TRUE \\
    reward\_iter & 30    & gen\_from\_policy & 10 \\
    irl\_episode\_per\_train & 10  &  term\_threshold & 100 \\
    meas\_samples & 5     & l1\_coeff & 0.1 \\
    gen\_samples & 10    &  n\_embed\_size & 32\\
    prop\_rounds & 2     &  net\_size & 256  \\
    \bottomrule
    \end{tabular}%
  \label{tab:hyperparam}%
\end{table*}%

\subsection{Metrics}
\label{app:metrics}
We provide several results on our constructed graphs and link predictions and below we discuss some information about the metrics reported in the tables and figures:

\begin{enumerate}
\item{Graph Construction Experiments}
We chose the following statistics that cover various aspects of graph structure:
\begin{itemize}
    \item Triangle Count: Number of triangles in the graph
    \item Clustering Coefficient: measure of the degree to which nodes in a graph tend to cluster together.
    \item Longest Connected Component: Size of the largest connected component.
    \item Assortativity: Pearson correlation of degrees of connected nodes, $A = \frac{{\rm cov}(X, Y)}{\sigma_X\sigma_Y}$ where the $(x_i
    , y_i)$ pairs are the degrees of connected nodes.
    \item Max Degree: Maximum degree of all nodes in the graph.
\end{itemize}

    \item Non-relational Link Prediction 
    Experiments
    \begin{itemize}
    \item AUC (Area under ROC Curve): Measures how well a parameter / model distinguish between correct and wrong connections. 
    \item Average Precision: Classical measure used to evaluate link prediction based on the precision recall performance of the model
    \end{itemize}
    
    \item Relational Link Prediction Experiments
    
    \begin{itemize}
    \item MRR (Mean Reciprocal Rank): This is a more robust statistic for evaluating link prediction compared to mean average rank and is widely used in relational learning tasks.
    \item HITS@K: It measures how many correct predictions made by the policy lie in the top $K$ predictions. 
    \end{itemize}
\end{enumerate}

Finally, we provide percent Deviation values with error in main paper in~\Cref{tab:gen_result}. This metric represent the percentage difference in graph bases statistics between the original graph and the generated graphs. As we generate multiple graphs to measure variation, we take the percent deviation with each of the generated graphs individually and then report the mean and error over these readings.
\newpage
\section{Additional Experiment Results}
\label{app:full-experiment}

\subsection{Synthesizing Graphs Via Learned Generative Mechanism}
\label{app:generative-capability}

In this section, we provide more comprehensive results on the construction capabilities of GraphOpt with results on two more datasets: Erdos-Renyi (Table~\ref{tab:ERgen}) and Pubmed (Table ~\ref{tab:Pubmedgen}). We also present results for the datasets we showed in the main paper along with these two as it contains two more statistics (Largest connected component and Assortativity) that we evaluated which we could not show in main paper due to space constraint. 

\begin{table*}[ht]
  \centering
  \caption{Percent deviation of graph statistics for generated graphs from observed BA graph (lower is better)}
    \resizebox{\textwidth}{!}{\begin{tabular}{cccccc}
    \toprule
    \textbf{Model} & \textbf{Triangle Count} & \textbf{Clustering Coeff.} & \textbf{Largest Connected Component} & \textbf{Assortativity} & \textbf{Max Degree} \\
    \toprule
    Observed Graph & 504 & 0.147 & 100 & -0.096 & 33 \\
    \midrule
    DC-SBM & $46.56 \pm 6.58$ & $59.44 \pm 7.11$ & $27.33 \pm 2.52$ & $91.61 \pm 0.64$ & $28.29 \pm 7.63$ \\
    BTER  & $48.02 \pm 9.11$  & $33.20 \pm 1.28$ & $35.33 \pm 0.58$ & $86.06 \pm 1.85$ & $33.33 \pm 0.00$ \\
    VGAE  & $70.89 \pm 8.95$ & $94.40 \pm 0.81$ & $9.00 \pm 1.00$ & $92.40 \pm 1.77$ & $8.08 \pm 1.75$ \\
    NetGAN & $31.68 \pm 6.28$ & $40.69 \pm 4.27$ & $\mathbf{4.00 \pm 1.73}$ & $62.12 \pm 35.49$ & $\mathbf{4.04 \pm 1.74}$ \\ 
    \bottomrule
    GraphOpt & $\mathbf{6.28 \pm 4.05}$ & $\mathbf{25.52 \pm 8.25}$ & $\mathbf{6.00 \pm 2.65}$ & $\mathbf{8.24 \pm 0.78}$ & $\mathbf{5.05 \pm 4.63}$ \\
    \bottomrule
    \end{tabular}}%
  \label{tab:BAgen}%
\end{table*}

\subsubsection{Erdos-Renyi Graph}

\begin{table*}[h!]
  \centering
    \caption{Percent deviation of graph statistics for generated graphs from the observed Erdos-Renyi graph (lower is better)}
    \resizebox{\textwidth}{!}{\begin{tabular}{cccccc}
    \toprule
    \textbf{Model} & \textbf{Triangle Count} & \textbf{Clustering Coeff.} & \textbf{Largest Connected Component} & \textbf{Assortativity} & \textbf{Max Degree} \\
    \toprule
    Observed Graph & 7335 & 4.84E-02  & 500 & -0.019 & 37  \\
    \midrule
    DC-SBM & $54.67 \pm 1.07$ & $94.91 \pm 0.74$ & $35.53 \pm 2.10$ & $70.17 \pm 9.36$ & $30.63 \pm 5.63$ \\
    BTER  & $60.11 \pm 0.85$ & $59.81 \pm 0.86$  & $38.87 \pm 2.48$ & $91.22 \pm 2.59$ & $8.11 \pm 4.68$  \\
    VGAE  & $85.62 \pm 0.44$ & $83.12 \pm 1.27$ & $\mathbf{2.33 \pm 0.31}$ & $82.81 \pm 11.05$ & $6.31 \pm 1.56$ \\
    NetGAN & $51.54 \pm 2.78$ & $49.47 \pm 5.89$  & $3.73 \pm 3.36$ & $\mathbf{31.58 \pm 22.94}$ & $\mathbf{2.73 \pm 2.70}$ \\
    \bottomrule
    GraphOpt & $\mathbf{26.29 \pm 2.89}$ & $\mathbf{23.96 \pm 3.68}$ & $6.07 \pm 1.22$ & $\mathbf{33.33 \pm 21.27}$ & $\mathbf{5.41 \pm 2.70}$ \\
    \bottomrule
    \end{tabular}}
  \label{tab:ERgen}
\end{table*}

\subsubsection{Political Blogs Graph}

\begin{table*}[h!]
  \centering
  \caption{Percent deviation of graph statistics for generated graphs from the observed Political Blogs Graph (lower is better)}
    \resizebox{\textwidth}{!}{\begin{tabular}{cccccc}
    \toprule
    \textbf{Model} & \textbf{Triangle Count} & \textbf{Clustering Coeff.} & \textbf{Largest Connected Component} & \textbf{Assortativity} & \textbf{Max Degree} \\
    \toprule
    Observed Graph & 303129 & 0.320 & 1222 & -0.221 & 351  \\
    \midrule
    DC-SBM & $52.78 \pm 9.15$ & $91.73 \pm 1.18$ & $65.25 \pm 10.79$ & $88.49 \pm 4.93$ & $40.86 \pm 1.89$ \\
    BTER  & $45.47 \pm 7.25$ & $54.17 \pm 13.57$ & $55.46 \pm 3.73$ & $90.75 \pm 4.83$ & $43.87 \pm 0.75$ \\
    VGAE  & $98.56 \pm 0.44$ & $99.32 \pm 0.55$ & $42.01 \pm 5.78$ & $97.86 \pm 2.03$ & $44.06 \pm 0.92$ \\
    NetGAN & $44.28 \pm 8.27$ & $37.55 \pm 7.20$ & $29.59 \pm 5.00$ & $\mathbf{24.95 \pm 13.79}$ & $\mathbf{38.75 \pm 3.70}$ \\
    \bottomrule
    GraphOpt & $\mathbf{34.73 \pm 3.79}$ & $\mathbf{20.34 \pm 9.12}$ & $\mathbf{24.30 \pm 2.05}$ & $\mathbf{21.79 \pm 4.78}$ & $\mathbf{36.85 \pm 2.71}$ \\
    \bottomrule
    \end{tabular}}
  \label{tab:PBgen}
\end{table*}

\newpage
\subsubsection{CORA-ML Graph}

\begin{table*}[h!]
  \centering
  \caption{Percent deviation of graph statistics for generated graphs from the observed Cora-ML graph}
    \resizebox{\textwidth}{!}{\begin{tabular}{cccccc}
    \toprule
    \textbf{Model} & \textbf{Triangle Count} & \textbf{Clustering Coeff.} & \textbf{Largest Connected Component} & \textbf{Assortativity} & \textbf{Max Degree} \\
    \toprule
    Observed Graph & 4890 & 0.241 & 2485 & -0.066 & 168 \\
    \midrule
    DC-SBM & $71.17 \pm 1.53$ & $68.25 \pm 20.16$ & $1.76 \pm 2.21$ & $26.04 \pm 7.49$ & $6.94 \pm 5.40$ \\
    BTER  & $40.06 \pm 1.17$ & $81.66 \pm 1.74$ & $5.15 \pm 2.85$ & $89.96 \pm 6.37$ & $16.47 \pm 14.49$ \\ 
    VGAE  & $99.56 \pm 0.24$ & $93.10 \pm 2.11$ & $\mathbf{0.05 \pm 0.02}$ & $96.86 \pm 1.52$ & $94.44 \pm 1.82$ \\ 
    NetGAN & $64.19 \pm 2.15$ & $41.12 \pm 18.82$ & $0.52 \pm 0.11$ & $\mathbf{2.84 \pm 3.16}$ & $4.17 \pm 2.38$ \\
    \bottomrule
    GraphOpt & $\mathbf{19.46 \pm 1.01}$ & $\mathbf{14.63 \pm 5.78}$ & $2.09 \pm 0.34$ & $\mathbf{4.10 \pm 3.53}$ & $\mathbf{2.58 \pm 1.24}$ \\
    \bottomrule
    \end{tabular}}
  \label{tab:Coragen}
\end{table*}

\subsubsection{PubMed Graph}

\begin{table*}[h!]
  \centering
  \caption{Percent deviation of graph statistics for generated graphs from the observed Pubmed graph}
    \resizebox{\textwidth}{!}{\begin{tabular}{cccccc}
    \toprule
    \textbf{Model} & \textbf{Triangle Count} & \textbf{Clustering Coeff.} & \textbf{Largest Connected Component} & \textbf{Assortativity} & \textbf{Max Degree} \\
    \toprule
    Observed Graph & 37560 & 6.02E-02 & 19717 & -0.0436 & 171  \\
    \midrule
    DC-SBM & $40.56 \pm 5.33$ & $85.33 \pm 8.31$ & $38.22 \pm 3.77$ & $83.87 \pm 18.17$ & $39.57 \pm 3.43$ \\
    BTER  & $35.17 \pm 4.69$ & $65.17 \pm 2.00$ & $41.74 \pm 2.34$ & $69.57 \pm 7.62$ & $37.43 \pm 5.064$ \\
    VGAE  & N/A & N/A & N/A & N/A & N/A \\
    NetGAN & $23.39 \pm 3.79$ & $53.32 \pm 7.13$ & $29.99 \pm 3.16$ & $57.57 \pm 12.76$ & $40.94 \pm 2.03$ \\
    \bottomrule
    GraphOpt & $\mathbf{19.95 \pm 2.30}$ & $\mathbf{42.47 \pm 2.33}$ & $\mathbf{13.63 \pm 1.39}$ & $\mathbf{18.19 \pm 11.87}$ & $\mathbf{23.19 \pm 6.00}$ \\
    \bottomrule
    \end{tabular}}%
  \label{tab:Pubmedgen}%
\end{table*}

\end{document}